\documentclass{article}
\usepackage{arxiv}

\usepackage[T1]{fontenc}
\usepackage[utf8]{inputenc}
\usepackage{amsmath,amssymb,mathtools}
\usepackage{booktabs}
\usepackage{graphicx}
\usepackage{xcolor}
\usepackage{microtype}
\usepackage{lineno}
\usepackage[numbers,square,sort&compress]{natbib}
\usepackage[colorlinks=true,
            linkcolor=blue!60!black,
            citecolor=green!55!black,
            urlcolor=blue!60!black]{hyperref}

\newcommand{\undertitle}{}
\newcommand{\TRE}{\textnormal{TRE}}
\newcommand{\PSNR}{\textnormal{PSNR}}
\newcommand{\MAE}{\textnormal{MAE}}
\newcommand{\NCC}{\textnormal{NCC}}
\newcommand{\SSIM}{\textnormal{SSIM}}
\newcommand{\LPIPS}{\textnormal{LPIPS}}
\newcommand{\AUSE}{\textnormal{AUSE}}
\newcommand{\reg}[1]{\texttt{#1}}

\title{What neurosurgeons need to see: synthetic intra-operative MRI from ultrasound for brain-shift compensation in brain tumour surgery}

\author{%
  Santiago Cepeda\,\textsuperscript{$\dagger$}\thanks{Corresponding author: \texttt{scepedac@saludcastillayleon.es}} \quad
  Olga Esteban-Sinovas\,\textsuperscript{$\dagger$} \quad
  Ignacio Arrese \quad
  Rosario Sarabia \\[0.6em]
  \normalsize Department of Neurosurgery, Neurovascular Unit \\
  \normalsize R\'io Hortega University Hospital, Valladolid, Spain \\[0.25em]
  \normalsize Specialized Group in Biomedical Imaging and Computational Analysis (GEIBAC) \\
  \normalsize Instituto de Investigaci\'on Biosanitaria de Valladolid (IBioVALL), Valladolid, Spain \\[0.5em]
  \normalsize \textsuperscript{$\dagger$}\textit{These authors contributed equally to this work.}
}

\renewcommand{\shorttitle}{Synthetic intra-operative MRI from ultrasound}

\providecommand{\headeright}{}
\begin{document}
\maketitle
\begin{abstract}
Maximal safe resection is the primary objective in glioma surgery. Neuronavigation provides useful intraoperative guidance, but its accuracy is progressively degraded by brain shift after dural opening. Intraoperative magnetic resonance imaging (MRI) can compensate for this deformation but requires dedicated infrastructure and is not widely available, whereas intraoperative ultrasound (ioUS) is inexpensive, repeatable, and compatible with routine surgical workflows. Navigation systems combining ioUS with preoperative MRI commonly rely on rigid registration; even deformable multimodal registration remains constrained by ultrasound speckle contrast, the limited field of view, and the inability to represent structures absent from the preoperative scan, most critically the resection cavity and residual tumor. We propose an end-to-end pipeline that generates a new whole-brain MRI volume in the preoperative imaging space by merging the preoperative MRI, a synthetic MRI generated from the ioUS, and a deformable registration anchored on that synthetic image. The pipeline integrates a 2.5D residual-transformer synthesis backbone (ResViT-2.5D), a two-stage registration coupling NiftyReg with a synthesis-anchored SynthMorph stage, and operates directly on raw scanner inputs. On a post-resection ReMIND cohort, ResViT-2.5D produced synthetic images closely matching the intraoperative T2 across structural, intensity, and perceptual similarity metrics. In 14 subjects with 215 expert landmarks, the synthesis-anchored registration reduced the mean target registration error from $6.27$ to $5.86$~mm, matching a strong classical NiftyReg baseline ($5.85$~mm)  while yielding a diffeomorphic deformation field in every subject. The contribution of this work is therefore not a gain in registration accuracy but the integrated whole-brain volume itself: inside the ultrasound field of view the volume reflects the intraoperative post-resection state, including the resection cavity, and outside it preserves the patient's preoperative MRI. This provides the surgeon with an MRI-like update of the operative field that registration alone cannot represent, with potential for integration into surgical-navigation workflows.
\end{abstract}

\keywords{Brain-shift compensation \and Intra-operative ultrasound \and Cross-modality image synthesis \and Deformable registration \and Surgical guidance \and ReMIND}

%% ============================================================================
\section{Introduction}\label{sec:intro}
%% ============================================================================
In glioma surgery, the operative goal is \textbf{maximal safe resection}: the removal of as much tumour tissue as possible while preserving neurological function \citep{herveyjumper_maximizing_2016}. The extent of resection is the strongest modifiable prognostic factor for survival \citep{sanai_extent_2008}, which makes the precise intraoperative localisation of the tumour boundary a central concern. The procedure is planned on a high-resolution preoperative MRI, but the brain exposed once the dura is opened is no longer the brain that was imaged. Cerebrospinal-fluid drainage, gravity-induced sag, tissue removal and cortical manipulation produce a continuous, time-varying soft-tissue deformation known as \textbf{brain shift} \citep{gerard_brain_2017}, which develops and accumulates over the course of the procedure \citep{nabavi_serial_2001}, reaching up to 20 to 25~mm at the cortical surface and several millimetres in deeper structures \citep{gerard_brain_2017}. From the moment of dural opening, the preoperative MRI displayed on the neuronavigation system becomes increasingly misaligned with the actual anatomy.

Intraoperative imaging is the established way to compensate for this. Intraoperative MRI (iMRI) provides the most informative update, but it requires dedicated infrastructure, prolongs the operation, and is available in only a small number of centres. Intraoperative ultrasound (ioUS) is the practical alternative: it is fast, inexpensive, repeatable, acquired with a tracked free-hand probe in any operating room, and able to delineate gliomas during surgery \citep{unsgaard_ability_2005,shetty_navigated_2021}. Its limitation is informational rather than logistical. An ioUS volume images only the tissue inside the surgical opening, in a speckle contrast that differs qualitatively from the MRI contrast on which the operation was planned and which the surgical team is trained to interpret. Beyond the ultrasound field of view, the preoperative MRI remains the only available signal and must be reconciled with the deformed intraoperative anatomy at the field-of-view boundary.

The standard computational tool for that reconciliation is deformable MRI-to-ioUS registration: the estimation of a non-rigid transformation that maps the preoperative MRI onto the intraoperative ioUS within a shared physical frame. Two decades of work have produced modality-invariant similarity metrics \citep{heinrich_mind_2013,hutchison_global_2013}, robust classical pipelines \citep{machado_deformable_2019}, learning-based displacement-field networks \citep{hoffmann_synthmorph_2022,wang_unsupervised_nodate}, and public benchmarks \citep{xiao_evaluation_2020,dorent_brain_2025} that place the achievable landmark target registration error (TRE) at approximately 2 to 5~mm on glioma-resection data.

Registration alone, however, cannot show the surgeon what the intraoperative anatomy looks like where the ioUS provides coverage but the preoperative MRI no longer corresponds, most importantly around the resection cavity. A warped preoperative MRI is, by construction, a deformed copy of preoperative tissue: the cavity that forms during the procedure has no counterpart voxels into which it could be deformed. This motivates a different deliverable, which we term a \textbf{synthetic intra-operative MRI} (siMRI): a whole-brain, MRI volume of the intraoperative state in which the region inside the ultrasound field of view is generated from the ioUS by a learned cross-modality synthesis, and the surrounding anatomy is supplied by the registered preoperative MRI.

Building such a pipeline raises three questions that have so far been studied in isolation. First, \emph{which synthesis architecture is suitable for clinical deployment?} Peak image quality at the training spacing does not guarantee robustness to the acquisition-spacing variability introduced by different vendors, probes and protocols in routine use. Second, \emph{does the synthesized MRI actually improve MRI-to-ioUS registration?} The published evidence is mixed, and we provide a controlled answer. Third, \emph{how should the synthesis and registration outputs be composed into a single whole-brain volume defined on the patient's preoperative imaging space?}

\subsection{Contributions}\label{sec:contributions}

This manuscript proposes a complete pipeline that answers the three questions above and reports its evaluation on the ReMIND dataset \citep{juvekar_remind_2024} with expert-placed landmark fiducials. Its contributions are the following.

\begin{enumerate}
\item \textbf{A backbone-selection study for deployment-ready ultrasound-to-MRI synthesis.} We compare three candidate backbones (ResViT-2.5D, ResViT-3D and SynDiff-2.5D, drawn from a companion cross-family benchmark on the same cohort \citep{estebansinovas_inprep}) against the intraoperative T2, and assess their robustness to acquisition-spacing shift and their inference cost, criteria that jointly determine clinical deployability rather than peak quality alone. On these criteria we adopt the 2.5D residual-transformer variant.

\item \textbf{Controlled, per-subject evidence that ultrasound-to-MRI synthesis does not improve deformable MRI-to-ioUS registration on post-resection data, with field-plausibility measured on the full deployed transformation.} We compose a NiftyReg \reg{reg\_aladin}~+~\reg{reg\_f3d} (MIND-SSC) stage with a synthesis-anchored SynthMorph residual and use this two-stage formulation as a controlled test of the recurring claim that a synthesized MRI improves MRI-to-ioUS registration; the synthesis-anchored pipeline matches a strong classical NiftyReg baseline on landmark target registration error rather than improving upon it. The role of the synthesis-anchored stage is instead to produce the MRI-versus-MRI-anchored warped preoperative MRI that serves as the canvas of the siMRI composition. We assess the geometric plausibility of the deformation, the fraction of voxels with non-positive Jacobian determinant (folding) and the smoothness of the log-Jacobian determinant (SDLogJ), on the \emph{composed} transformation actually applied to the preoperative MRI rather than on each stage in isolation, since a stage that is fold-free on its own may still introduce folding once composed with the others (Section~\ref{sec:plausibility}).

\item \textbf{Synthetic intraoperative MRI (siMRI) composition delivering a whole-brain volume in the patient's preoperative imaging space.} Inside the ultrasound field of view the siMRI is supplied by the synthesis output; outside it, by the registration-warped preoperative MRI, with a feathered transition at the cone boundary, producing the final volume in both the intraoperative cube and the preoperative space (Section~\ref{sec:pseudo-mr}).  To the best of our knowledge, this is the first reported end-to-end pipeline that delivers a whole-brain MRI volume in the patient's preoperative imaging space from a single intraoperative ioUS and the preoperative MRI.
\end{enumerate}

The rest of the manuscript is organised as follows. Section~\ref{sec:related} surveys the three lines of work that the pipeline integrates. Section~\ref{sec:methods} describes the data, the pipeline and the evaluation protocol. Section~\ref{sec:results} reports the three experimental blocks. Section~\ref{sec:discussion} discusses the implications, limitations and directions for clinical deployment.
%% ============================================================================
\section{Related work}\label{sec:related}
%% ============================================================================

Three lines of work intersect in the proposed pipeline: cross-modal MRI-ultrasound image synthesis, MRI-ioUS deformable registration for brain-shift correction, and the use of synthesised modality bridges to translate cross-modality registration into an intra-modal problem. We review each, focusing on works evaluated on glioma-resection data, and particularly on the ReMIND release.

\subsection{Cross-modal MRI-ultrasound synthesis}\label{sec:related-synth}

Synthesis has been pursued in both directions, each motivated by a distinct downstream goal.

\textbf{MRI$\,\to\,$US synthesis} is the natural choice when the goal is to generate ultrasound-like data for training US-domain tasks, most commonly segmentation. The strongest current evidence is provided by MMHVAE \citep{dorent_unified_2026}, which trains an nnU-Net exclusively on synthesised ioUS and reports a median Dice of 73.6 on the RESECT-SEG benchmark, matching a real-ioUS supervised baseline. CCLD \citep{jiang_cross-modal_2025} pursues the same direction with a latent-diffusion model and a frequency-decomposed feature-fusion module, reporting state-of-the-art MRI$\,\to\,$ioUS image-quality metrics on ReMIND. Outside the brain, Azampour et al. \citep{azampour_multitask_2024} use MRI$\,\to\,$TRUS synthesis as a weakly supervised bridge for prostate registration. A complementary physics-based route is taken by DiffUS \citep{bertramo_diffus_2025}, which maps T1 intensities to acoustic impedance and renders B-mode ultrasound by differentiable wave propagation. The common feature of these works is that the synthetic ultrasound is a \emph{training-data} or \emph{simulation} product; the surgeon is not expected to interpret it directly.

\textbf{US$\,\to\,$MRI synthesis} reverses the direction and targets the intraoperative workflow itself: the synthesised MRI is the image that the surgeon and the navigation pipeline can readily interpret, because the procedure was planned on MRI. The earliest contributions are in the fetal-brain domain \citep{jiao_self-supervised_2020,silverstein_translation_nodate}, the latter introducing DDIC, a dual-diffusion approach with an explicit cross-domain correlation loss. In the intraoperative brain setting, BrainVoxGen \citep{singh_brainvoxgen_2023} demonstrated 3D Pix2Pix synthesis on 18 ReMIND-style cases, with the authors themselves describing the results as below deployment thresholds. MHVAE and MMHVAE \citep{dorent_mhvae_2023,dorent_unified_2026} support both directions through a single hierarchical multimodal VAE and, in the ioUS$\,\to\,$MRI direction, match conditional-GAN baselines (ResViT, Pix2Pix) on PSNR, SSIM and LPIPS. In other anatomies, Salmanpour et al. \citep{salmanpour_influence_nodate} benchmark ten image-to-image networks on prostate US$\,\to\,$MRI translation and show that pixel-similarity metrics do not capture downstream clinical fidelity. The present manuscript belongs to this US$\,\to\,$MRI group: we deploy the residual-transformer hybrid of Dalmaz et al. \citep{dalmaz_resvit_2022} (ResViT) in a 2.5D inference regime, selected from a companion cross-family benchmark on the ReMIND cohort \citep{estebansinovas_inprep}. Two further backbones are retained for analysis: the same architecture as a full-3D generator, and the few-step adversarial diffusion model SynDiff \citep{ozbey_syndiff_2023}, adapted to a single-direction paired formulation.

\subsection{MRI-ioUS deformable registration for brain-shift correction}\label{sec:related-reg}

The classical line uses modality-invariant similarity metrics with free-form deformation. Wein et al. \citep{hutchison_global_2013} introduced the LC$^2$ similarity for rigid US-to-MRI alignment, still the standard rigid pre-processing step. Heinrich et al. \citep{heinrich_mind_2013} proposed MIND-SSC, a local self-similarity descriptor that is by construction invariant to monotonic intensity transformations and that became the dominant similarity term for MRI-ioUS deformable registration. The most comprehensive multi-site evaluation is that of Machado et al. \citep{machado_deformable_2019} with cDRAMMS, reaching a mean $\TRE$ of 2.08 to 2.28~mm across BITE, RESECT and MIBS with a single fixed parameter set. The RESECT benchmark and its first public head-to-head registration challenge (CuRIOUS) are described in Xiao et al. \citep{xiao_evaluation_2020}; on its private test set, classical methods outperformed the only deep-learning entry, which overfitted strongly.

Public benchmarking has since been consolidated under the Learn2Reg ReMIND2Reg series \citep{dorent_brain_2025,hansen_learn2reg_2025}. On the 2024 leaderboard, the classical NiftyReg baseline retained first place (mean $\TRE$ 2.87~mm) ahead of every learning-based entry, illustrating that modality-invariant descriptors combined with block matching remain competitive on the post-resection task. The current top performer on the 2025 edition is MCPO \citep{wang_unsupervised_2026}, an extension of MCBO \citep{wang_unsupervised_nodate} that combines ConvexAdam \citep{siebert_convexadam_2025} with multilevel correlation pyramids and a patch-based mutual-information instance-optimisation refinement. A distinct paradigm is introduced by Morozov et al. \citep{morozov_3d_2025}, with patient-specific 3D keypoint descriptors trained by matching-by-synthesis: synthetic ioUS volumes generated from each patient's MRI by MMHVAE provide the cross-modal training pairs for a Siamese 3D ResNet-18 descriptor.

\subsection{Synthesis-bridged registration and positioning}\label{sec:related-bridge}

A recurring idea is that inter-modality matching can be reduced to intra-modality matching by synthesising one modality from the other. Variants in the literature include synthesising MRI from ioUS and matching against the real MRI \citep{wang_unsupervised_nodate,wang_unsupervised_2026,jiang_cross-modal_2025}, synthesising ioUS from MRI for keypoint training \citep{morozov_3d_2025}, and dual-channel formulations \citep{wang_unsupervised_2026}. The empirical evidence on registration accuracy is mixed. On ReMIND2Reg, Wang et al. \citep{wang_unsupervised_nodate} report a modest improvement from the synthesis bridge that is not statistically significant on a per-subject basis. On simulated rigid deformations within MMHVAE \citep{dorent_unified_2026}, the synthesised modality enables registration to recover translations that the raw ioUS-MRI pair cannot, but the demonstration is restricted to rigid transformations.

Our development experiments (Section~\ref{sec:results-ablation}) reach a complementary conclusion: a modality-invariant descriptor (MIND-SSC) already saturates the classical signal, so substituting the synthetic MRI for the ioUS as the fixed image of the classical solver does not measurably help. The synthetic MRI does, however, provide a useful anchor for a learning-based displacement-field network \citep[SynthMorph;][]{hoffmann_synthmorph_2022}, whose training prior covers only MRI sequences and which produces small-magnitude displacements when applied directly to the raw (ioUS, MRI) pair. The present manuscript integrates these three lines into an end-to-end pipeline that delivers a whole-brain synthetic T2-weighted MRI aligned with the patient's preoperative imaging space, with the synthesis backbone selected on its quality against the intraoperative T2 and its operational suitability for deployment. To the best of our knowledge, no prior work has reported an end-to-end pipeline that yields a synthetic T2-weighted MRI in the patient's preoperative imaging space from a single intraoperative ioUS and a single preoperative MRI.

%% ============================================================================
\section{Materials and methods}\label{sec:methods}
%% ============================================================================

\begin{figure}[t]
  \centering
  \includegraphics[width=\linewidth]{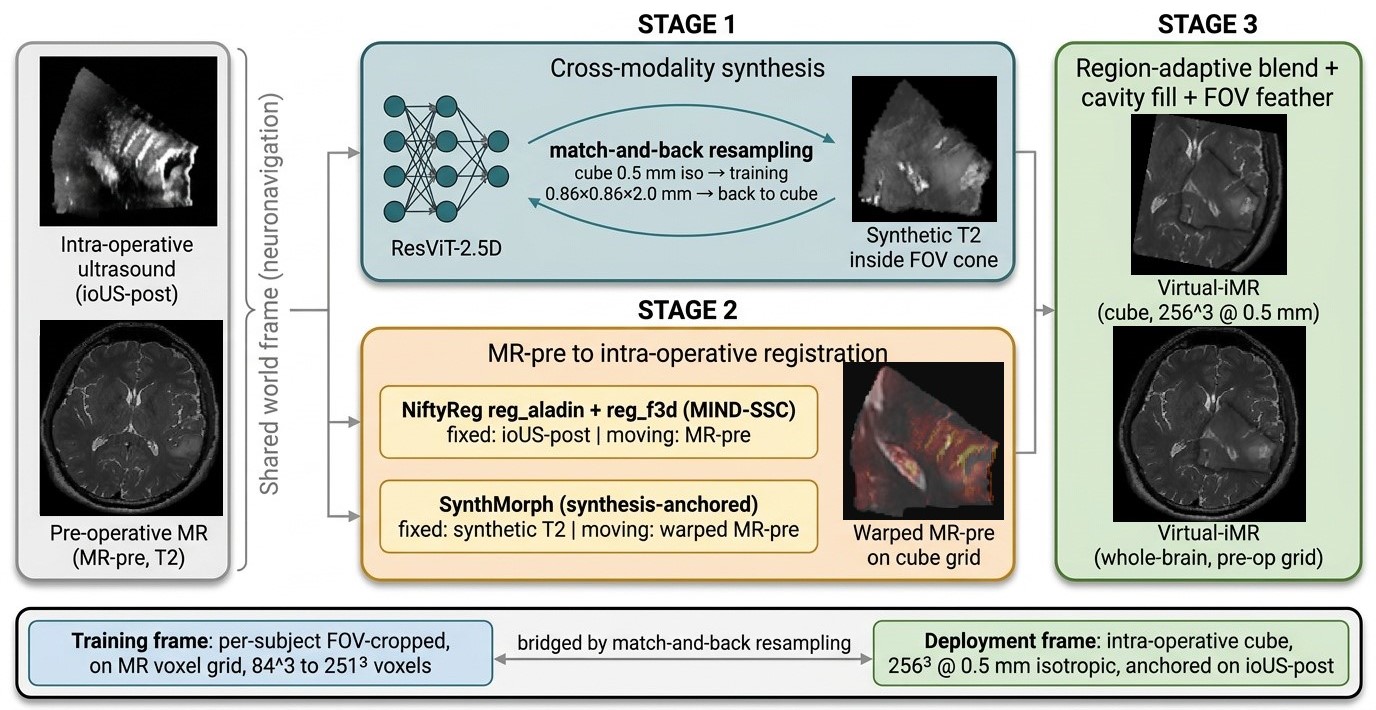}
  \caption{Overview of the proposed end-to-end pipeline. From a single  post-resection intraoperative ultrasound (ioUS) and the patient's  preoperative T2-weighted MRI,  both sharing the neuronavigator world frame, the pipeline produces a whole-brain synthetic intraoperative MRI (siMRI) aligned with the preoperative imaging space, in three stages.
  \textbf{Stage 1}: cross-modality synthesis with ResViT-2.5D, applied inside the cube via a match-and-back resampling protocol that bridges the cube spacing (0.5~mm isotropic) and the network training spacing ($0.86 \times 0.86 \times 2.0$~mm). \textbf{Stage 2}: registration of preoperative MRI onto the intraoperative cube, combining NiftyReg (\reg{reg\_aladin} + \reg{reg\_f3d} with MIND-SSC similarity) with a synthesis-anchored SynthMorph residual. \textbf{Stage 3}: composition of the synthesis content (inside the FOV cone) with the warped preoperative MRI (outside it) through a feathered FOV transition; the deliverable is output at the cube resolution and back-projected to the patient's preoperative imaging space. The two reference frames involved, the per-subject FOV-cropped training frame and the per-subject $256^{3}$ intraoperative deployment cube, are connected only at inference time by the match-and-back resampling.}
  \label{fig:pipeline}
\end{figure}

The end-to-end pipeline is summarised in Figure~\ref{fig:pipeline}: a single intraoperative ultrasound feeds a cross-modality synthesis stage, whose output anchors a two-stage registration of the preoperative MRI onto the intraoperative cube; a composition step delivers the whole-brain synthetic intraoperative MRI.

\subsection{Cohort curation and splits}\label{sec:splits}

All experiments were performed on the publicly released ReMIND cohort \citep{juvekar_remind_2024}, the Brain Resection Multimodal Imaging Database (Brigham and Women's Hospital, Boston, MA, USA), which distributes, per subject, intraoperative 3D ultrasound (ioUS) acquired during brain-tumour resection, preoperative MRI (T2-weighted, T2-FLAIR when available, contrast-enhanced T1-weighted), and intraoperative MRI acquired after partial or complete resection. For each subject, ReMIND distributes two ioUS acquisition phases: a pre-resection acquisition through the dural opening and a post-resection acquisition at the end of resection. The present pipeline operates on the post-resection ioUS, which is the acquisition closest in time to the intraoperative MRI and the most clinically relevant for navigation after tissue removal.

\paragraph{Choice of T2-weighted as synthesis and registration target.} Two reasons motivate the choice of T2-weighted contrast. First, T2-weighted is the only contrast available at every paired time point in the cohort: across the 76-subject curated working cohort, 153 paired ioUS/T2w studies span pre- and post-resection acquisitions, whereas only 104 paired ioUS/FLAIR studies meet the same pairing criterion and the ioUS/T1ce coverage is comparable. The shortfall is concentrated on the post-resection acquisitions, in which the operating-theatre protocol is compressed and the additional sequences are frequently skipped. Second, and more fundamentally, the ReMIND cohort spans a heterogeneous set of histologies, including high-grade gliomas, low-grade gliomas and other pathologies. The appearance of these tumours on preoperative contrast-enhanced T1-weighted (T1ce) is highly variable, with high-grade gliomas typically showing avid enhancement and low-grade gliomas remaining non-enhancing; combined with the fact that the intraoperative MRI in ReMIND is acquired without contrast, this would force a hypothetical T1ce synthesis network to learn a mapping whose target contrast varies dramatically from subject to subject. T2-weighted contrast is comparatively uniform across pathologies and acquisition stages, which makes it the most learnable target for cross-modality synthesis on this cohort.

\paragraph{Curation and partition.} The 76-subject working cohort was curated from the public ReMIND release by visual quality control on the ioUS volumes. Studies were excluded when the ultrasound acquisition was unusable for paired learning: insufficient acoustic coupling, large shadowing artefacts, near-empty foreground inside the dural opening, or marked probe-induced geometric distortion that compromised the rigid co-registration to the preoperative MRI. The curated cohort was divided into two non-overlapping subject-level partitions. The \textbf{synthesis training partition} contains 60 subjects (122 paired ioUS/T2w studies) and was used to fit the synthesis backbone; the deployed checkpoint of the present manuscript is the ResViT-2.5D model selected by the criteria of Section~\ref{sec:backbone-selection}, re-used here without retraining or fine-tuning. The \textbf{held-out partition} contains 16 subjects (31 paired studies), and the \textbf{14-subject primary test cohort} on which the registration and siMRI experiments of Section~\ref{sec:results} operate is a strict subset of this held-out partition. No ioUS or MRI pair seen at evaluation time was ever seen by the synthesis network at training time. The pre-resection and post-resection studies of a given patient were assigned to the same partition, so intra-subject leakage is prevented by construction. The partition was sampled once with a fixed pseudo-random seed (NumPy \texttt{default\_rng(seed=42)}) on the alphabetically ordered list of curated subject identifiers, and is reused verbatim by every synthesis backbone evaluated in this work. The 21\%~/~79\% (16~/~60) split ratio was chosen as a compromise between training-set size and the availability of a sufficiently large held-out partition to support per-subject paired statistical testing; no stratification by tumour grade, anatomical location or ioUS acquisition vendor was applied.

\paragraph{Landmark protocol.} For the 14 subjects of the primary test cohort, anatomical landmark pairs were placed in both the preoperative MRI and the post-resection ioUS following the annotation protocol of the ReMIND2Reg challenge \citep{dorent_brain_2025}. Two expert observers identified corresponding landmarks using the markups tool of 3D Slicer (v5.10.0; \citealp{fedorov_slicer_2012}): the two annotators selected landmarks alternately, each proposal was reviewed and, where necessary, adjusted by the other observer until consensus was reached, and on completion both observers jointly reviewed every landmark and discarded those lacking unanimous agreement. We extended the eligible landmark set beyond the deep sulcal grooves, convex gyral points and sulcal vanishing points used by ReMIND2Reg to also cover peri-lesional and intervention-relevant structures, namely (a) vessel bifurcations, (b) cortical-surface landmarks, (c) ventricular margins and choroid plexus, (d) tumour and resection-cavity boundaries, and (e) corresponding points between residual tumour and preoperative tumour. This extension deliberately samples the peri-cavity and tumour-boundary regions that are the hardest to register and the most relevant to the surgical objective. The full set comprises 215 valid landmark pairs across the 14 subjects (median 14 pairs per subject, range 12 to 22). For this class of landmarks, an inter-observer localisation variability of approximately $1.89$~mm has been reported in 3D intraoperative ultrasound \citep{machado_nonrigid_2018} and is adopted as the reference annotation floor of the ReMIND2Reg benchmark \citep{dorent_brain_2025}. This figure is a conservative lower bound for the present set, since our landmarks are localised across the preoperative MRI and the post-resection ioUS rather than within ultrasound, the annotation involves an additional transfer to the MRI space, and our eligible sites additionally include the harder peri-cavity and tumour boundaries.

\subsection{Cross-modality MRI synthesis from intraoperative ultrasound}\label{sec:synth}

The synthesis backbone was selected from a prior benchmark we conducted on the same ReMIND cohort \citep{estebansinovas_inprep}, in which three architectural families (convolutional GAN baselines, the residual-transformer hybrid ResViT, and the few-step adversarial-diffusion model SynDiff) were evaluated under four inference regimes (2D, 2.5D, 2D~+~3D-refinement, and full-3D). Among the configurations tested there, ResViT-2.5D \citep{dalmaz_resvit_2022} offered the best balance of perceptual quality and operational suitability for deployment, and is the configuration deployed in the present pipeline. ResViT-3D and SynDiff-2.5D \citep{ozbey_syndiff_2023} are kept as reference points throughout this manuscript.

\subsubsection{Architectural families}\label{sec:synth-arch}

\begin{figure}[!htbp]
  \centering
  \includegraphics[width=0.8\linewidth]{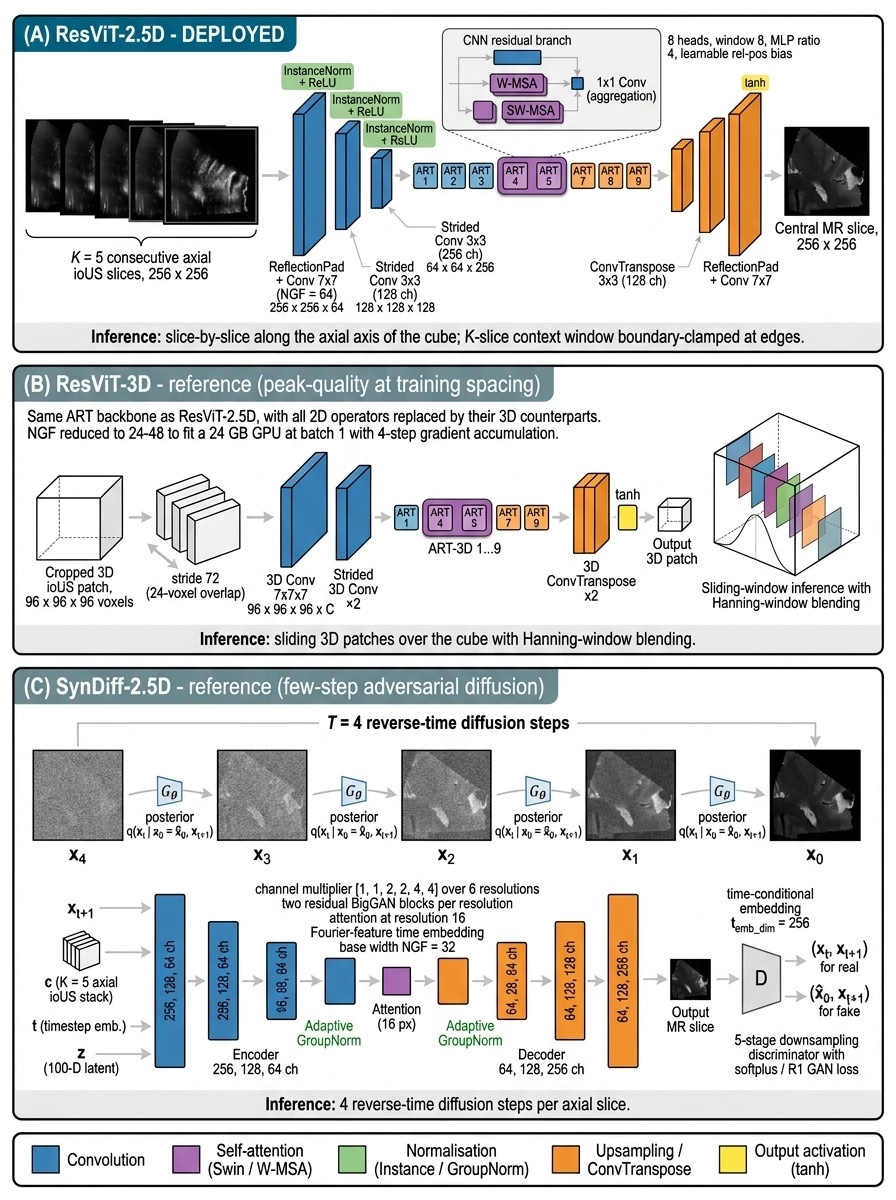}
  \caption{Architectures of the three candidate synthesis backbones.
  \textbf{(A) ResViT-2.5D, the deployed model:} an encoder--decoder with
  aggregated residual transformer (ART) blocks, two of which carry a parallel
  Swin-Transformer branch; it maps a $K = 5$ stack of consecutive axial ioUS
  slices ($256 \times 256$) to the central MRI slice, inferred slice-by-slice
  along the cube axial axis. \textbf{(B) ResViT-3D, peak-quality reference:}
  the same ART backbone with all operators in 3D, run on $96^{3}$ patches with
  overlap blending. \textbf{(C) SynDiff-2.5D, adversarial-diffusion reference:}
  a few-step ($T = 4$) adversarial diffusion model conditioned on the same
  $K = 5$ ioUS stack. Hyperparameters are detailed in Section~\ref{sec:synth-arch}.
  The colour legend at the bottom is shared across panels.}
  \label{fig:architectures}
\end{figure}

The three candidate synthesis backbones are summarised in Figure~\ref{fig:architectures}; their key design choices are described below.

\textbf{ResViT-2.5D and ResViT-3D.} ResViT \citep{dalmaz_resvit_2022} is a residual-transformer hybrid generator that combines a fully convolutional encoder--decoder with a self-attention bottleneck. The bottleneck contains a stack of nine aggregated residual transformer (ART) blocks; the two blocks at positions four and five carry a parallel shifted-window Swin-Transformer branch, whose feature map is concatenated with the convolutional branch and aggregated by a $1 \times 1$ convolution. The remaining seven blocks are convolution-only residual blocks. The encoder is a reflection-padded $7 \times 7$ convolution followed by two strided $3 \times 3$ convolutions with instance normalisation and ReLU; the decoder mirrors the encoder through transposed convolutions and ends in a $\tanh$ activation. The discriminator is a three-stage strided PatchGAN with $4 \times 4$ kernels and LeakyReLU activations.

We deploy ResViT under a \textbf{2.5D inference regime}: the input is a stack of $K = 5$ consecutive axial ioUS slices and the target is the central MRI slice. At inference the generator runs slice by slice along the axial axis of the cube; the $K$-slice context window is boundary-clamped at the edges of the volume. The 2.5D regime trades the strict volumetric continuity of a full-3D variant for inference at a 2D parameter budget and for an inference cost compatible with real-time, streaming deployment; its robustness to acquisition-spacing shift is comparable to that of the full-3D variant (Section~\ref{sec:results-spacing}), which matters in a clinical-deployment setting in which the ioUS spacing is not under the user's control.

\textbf{ResViT-3D} replaces every 2D operator in the generator and discriminator by its 3D counterpart, operating on $96^3$ patches with 24-voxel overlap and Hanning-window blending at inference; the generator width is reduced (NGF~=~24--48 depending on the VRAM budget) to fit a 24~GB GPU at batch size 1 with four-step gradient accumulation. The ART blocks remain at the same depth, with the Swin-Transformer branch operating on 3D volumetric windows.

\textbf{SynDiff-2.5D.} SynDiff \citep{ozbey_syndiff_2023} is an adversarial diffusion model built on the DDGAN framework. We adapted the original bidirectional cycle-consistent formulation to a single-direction paired setting: only the diffusive generator and its discriminator are retained, conditioned directly on the paired ioUS. The generator is an NCSNpp time-conditional U-Net with adaptive group normalisation and Fourier-feature time embedding (channel multiplier $[1, 1, 2, 2, 4, 4]$ over six resolutions, two residual blocks per resolution, BigGAN-style residual blocks, attention at the 16-pixel resolution, base width NGF~=~32 for 2D and 2.5D). The diffusion process is variance-preserving SDE with $\beta_{\min} = 0.1$, $\beta_{\max} = 20$ and only $T = 4$ reverse-time steps (the DDGAN few-step regime). At each step the generator receives $(x_{t+1}, c, t, z)$ with $c$ the conditioning ioUS and $z \sim \mathcal N(0, I)$ a 100-dimensional latent, and predicts $\hat x_0$. The next-step sample is drawn from the analytical posterior $q(x_t \mid x_0 = \hat x_0, x_{t+1})$. The 2.5D regime inputs a stack of $K = 5$ axial ioUS slices alongside the noisy MRI target.

\subsubsection{Training procedure}\label{sec:synth-train}

The pre-processing of every (ioUS, MRI) study follows the protocol of our prior synthesis benchmark and is summarised here for completeness. Each pair is rigidly co-registered with the LC$^2$ metric \citep{hutchison_global_2013} in ImFusion Suite (ImFusion GmbH, Munich, Germany; v2.42.2), using the ioUS as reference and the MRI as moving image. After registration, the ioUS is resampled with B-spline interpolation onto the voxel grid of its MRI partner (preoperative MRI for the pre-resection pair, intraoperative MRI for the post-resection pair). A tight per-subject bounding box is then extracted from the ioUS foreground (intensity above 1\% of the volume maximum) and applied to both volumes, so that the training tensors inherit the FOV cone shape of the ioUS rather than a regular cube; the resulting cropped volumes range in size from 84$^3$ to 251$^3$ voxels and approximately match the surgical opening. For 2D and 2.5D training, axial slices are then resized in-plane to $256 \times 256$ (depth left at native size for the 2.5D channel stack); full-3D training operates on the native cropped volume size. Per-modality intensity normalisation follows Equations~\eqref{eq:us-norm} and \eqref{eq:mr-norm}.

Light data augmentation is applied at training time only: random horizontal and vertical flips ($p = 0.5$), random in-plane rotations of $\pm 15^\circ$ ($p = 0.3$), and ioUS-only intensity perturbation (gamma $\gamma \in [0.8, 1.2]$ together with additive Gaussian noise of standard deviation $\sigma \in [0, 0.03]$, $p \approx 0.4$).

The synthesis network is a \emph{single model trained jointly on the union of (pre-resection ioUS, preoperative MRI) and (post-resection ioUS, intraoperative MRI) paired studies}; no separate pre- or post-resection branches are instantiated. At deployment the same network ingests the post-resection ioUS that is the only ioUS available after the surgical opening; its target is therefore a synthetic intraoperative T2-weighted MRI.

The objective combines a least-squares adversarial term \citep{mao_lsgan_2017}, a strong voxel-wise $L_1$ supervision term, and a feature-matching term \citep{wang_pix2pixhd_2018} computed across all discriminator layers. For a paired sample $(\mathbf{x}, \mathbf{y})$ with $\mathbf{x}$ the ioUS context stack and $\mathbf{y}$ the central MRI slice, let $\hat{\mathbf{y}} = G(\mathbf{x})$ be the generator prediction and let $D^{(l)}(\cdot)$ denote the activations of the $l$-th discriminator layer. The discriminator $D$ minimises
\begin{equation}\label{eq:lsd}
\mathcal{L}_{D} \;=\; \tfrac{1}{2}\,\mathbb{E}_{(\mathbf{x},\mathbf{y})}\!\big[(D(\mathbf{x}, \mathbf{y}) - 1)^2\big] + \tfrac{1}{2}\,\mathbb{E}_{\mathbf{x}}\!\big[D(\mathbf{x}, G(\mathbf{x}))^2\big],
\end{equation}
and the generator $G$ minimises the combined objective
\begin{equation}\label{eq:lg}
\mathcal{L}_{G} \;=\; \lambda_{adv}\, \mathcal{L}_{G,adv} + \lambda_{L_1}\, \mathcal{L}_{L_1} + \lambda_{fm}\, \mathcal{L}_{fm},
\end{equation}
with
\begin{equation}\label{eq:lg-terms}
\mathcal{L}_{G,adv} = \tfrac{1}{2}\,\mathbb{E}_{\mathbf{x}}\!\big[(D(\mathbf{x}, G(\mathbf{x})) - 1)^2\big], \quad
\mathcal{L}_{L_1} = \mathbb{E}_{(\mathbf{x},\mathbf{y})}\!\big[\| G(\mathbf{x}) - \mathbf{y} \|_1 \big],
\end{equation}
\begin{equation}\label{eq:lfm}
\mathcal{L}_{fm} = \mathbb{E}_{(\mathbf{x},\mathbf{y})}\!\left[ \sum_{l=1}^{L} \tfrac{1}{N_l}\, \| D^{(l)}(\mathbf{x},\mathbf{y}) - D^{(l)}(\mathbf{x}, G(\mathbf{x})) \|_1 \right].
\end{equation}
We use $\lambda_{adv} = 1$, $\lambda_{L_1} = 100$, $\lambda_{fm} = 10$. Optimisation uses Adam ($\beta_1 = 0.5$, $\beta_2 = 0.999$) with a cosine annealing schedule from the base learning rate to $1 \times 10^{-6}$, gradient clip-norm 1.0, and bfloat16 autocast on a single 24~GB GPU.

Training proceeds in two phases. \textbf{Phase~1} trains the generator with the Swin-Transformer branch disabled (a convolution-only generator) for 100 epochs at a base learning rate of $2 \times 10^{-4}$. \textbf{Phase~2} loads the best Phase~1 checkpoint, enables the Swin branch, and continues training for 100 epochs at a base learning rate of $1 \times 10^{-4}$; the convolutional weights remain trainable so that the two branches can co-adapt. An exponential moving average copy of the generator (decay 0.999) is maintained throughout and is the version used at inference. For SynDiff-2.5D the analogous schedule is 300 epochs at $\text{lr}_G = 1.6 \times 10^{-4}$ and $\text{lr}_D = 1 \times 10^{-4}$, with the $R_1$ gradient penalty applied lazily every 64 steps and a softplus discriminator loss per timestep.

\subsubsection{Inference inside the cube: match-and-back resampling}\label{sec:synth-inference}

At deployment we need a synthetic MRI co-located with the ioUS in the intraoperative cube (Section~\ref{sec:cube}). The cube grid ($256^3$ at 0.5~mm isotropic) does not in general match the spacing of the ioUS volume after its rigid co-registration to the patient's MRI during training. We bridge the two frames at inference time with a \emph{match-and-back resampling} protocol applicable to each of the three backbones:

\begin{enumerate}
\item Start from $\text{ioUS}_{\text{cube}}$ at $256^3$ at 0.5~mm.
\item Resample to the training spacing (approximately $0.86 \times 0.86 \times 2.0$~mm in our cohort).
\item Run the synthesis network on the resampled grid (slice-by-slice for ResViT-2.5D and SynDiff-2.5D; sliding 3D windows with Hanning blending for ResViT-3D).
\item Resample the synthetic MRI back to the cube grid ($256^3$ at 0.5~mm).
\item Apply the FOV mask to obtain the input to the registration stage.
\end{enumerate}

The choice of intermediate spacing matches the dominant spacing in the synthesis training distribution (median approximately $0.86 \times 0.86 \times 2$~mm). The internal $256 \times 256$ slice resize of the ResViT-2.5D architecture is invariant to in-plane spacing by construction, so the only round-trip artefact for that backbone is the through-plane resampling (small under trilinear interpolation); ResViT-3D operates on voxel-native sliding 3D windows and is therefore sensitive to the entire spacing vector; SynDiff-2.5D inherits the same in-plane prior as ResViT-2.5D but adds a stochastic sampling step at every reverse-time iteration. Section~\ref{sec:results-spacing} quantifies the empirical robustness of the three backbones under this domain shift.

\subsection{Reference frames}\label{sec:cube}

The pipeline operates within two distinct reference frames whose relationship must be made explicit.

\textbf{Training frame.} The synthesis network is trained in the voxel grid of the MRI partner of each (ioUS, MRI) pair: the preoperative MRI grid for the pre-resection pairs and the intraoperative MRI grid for the post-resection pairs (Section~\ref{sec:synth-train}). Per-subject training tensors are variable in shape (84$^3$ to 251$^3$ voxels) and inherit the FOV cone shape of the ioUS rather than a regular cube.

\textbf{Deployment frame.} The registration and composition stages operate on a per-subject \textbf{intraoperative cube} of $256^3$ voxels at 0.5~mm isotropic spacing, covering a 12.8~cm field of view per axis. The cube affine is anchored on the post-resection ioUS volume, which is always present at inference and whose acquisition cone defines the operating field of view. Isotropic 0.5~mm sampling was chosen to avoid the $z$-axis aliasing introduced by the anisotropic native intraoperative MRI spacing (around 2~mm slice thickness) and to remain compatible with the in-plane resolution of high-end preoperative MRI. The four available volumes per subject (preoperative MRI, intraoperative MRI, pre- and post-resection ioUS) are placed on the cube grid by trilinear interpolation (nearest-neighbour for segmentations).

\textbf{Deployment inputs.} The pipeline operates on the \textbf{raw ReMIND NIfTIs} (post-resection ioUS and preoperative MRI), each carrying its own neuronavigator-derived affine and no preprocessing rigid co-registration of the modalities. The deployment cube is defined on the raw post-resection ioUS as described above; the synthesis network bridges the raw ioUS spacing to its training spacing at inference time through the match-and-back resampling protocol (Section~\ref{sec:synth-inference}). The full evaluation suite (Section~\ref{sec:results}) operates on this raw-input pipeline so that every reported $\TRE$, field-plausibility figure and pipeline wallclock reflects the operating-room scenario.

\paragraph{Intensity normalisation.} Let $\Omega_{fg}$ denote the foreground voxel set of a volume (intensity above 1\% of the volume maximum). The post-resection ioUS is robust-percentile rescaled and mapped to $[-1, 1]$:
\begin{equation}\label{eq:us-norm}
\tilde{x}_{\text{ioUS}}(\mathbf{r}) \;=\; 2 \cdot \frac{\operatorname{clip}\!\big(x_{\text{ioUS}}(\mathbf{r});\, p_2,\, p_{98}\big) - p_2}{p_{98} - p_2} - 1, \quad \mathbf{r} \in \Omega_{fg},
\end{equation}
with $p_2$ and $p_{98}$ the per-subject 2nd and 98th intensity percentiles on $\Omega_{fg}$; background voxels are set to $-1$. The MRI volumes are foreground z-scored and clipped at $\pm 3\sigma$ before linear remapping to $[-1, 1]$:
\begin{equation}\label{eq:mr-norm}
\tilde{x}_{\text{MR}}(\mathbf{r}) \;=\; \tfrac{1}{3}\,\operatorname{clip}\!\left(\frac{x_{\text{MR}}(\mathbf{r}) - \mu_{fg}}{\sigma_{fg}};\, -3,\, +3 \right), \quad \mathbf{r} \in \Omega_{fg},
\end{equation}
where $(\mu_{fg}, \sigma_{fg})$ are the foreground mean and standard deviation. The FOV mask is obtained from the ioUS cone after morphological closing $\mathcal{C}_k$ (kernel $k=2$) and hole filling $\mathcal{H}$:
\begin{equation}\label{eq:fov-mask}
\Omega_{\text{FOV}} \;=\; \mathcal{H}\!\left( \mathcal{C}_{k=2}\!\left( \mathbb{1}_{\{x_{\text{ioUS}}>0\}} \right) \right),
\end{equation}
and the FOV pre-masking of the moving image required for the rigid initialisation (Section~\ref{sec:rigid}) is the pointwise product $\tilde{x}_{\text{MR}}^{\Omega} = \tilde{x}_{\text{MR}} \cdot \mathbb{1}_{\Omega_{\text{FOV}}}$.

\subsection{Preoperative MRI to intraoperative registration}\label{sec:reg}

\subsubsection{Starting condition: world-frame alignment from neuronavigation}\label{sec:starting}

A clarification on the starting condition is necessary because it determines what registration is being asked to recover. The ReMIND volumes are exported from the neuronavigation workstation that the surgical team used to plan and guide the procedure; they are distributed in DICOM format and converted to NIfTI for the present pipeline. Each modality is thereby associated with an affine matrix encoding the patient-specific transformation from voxel to a \textbf{shared physical world frame}, the same world frame used by the neuronavigator to track the patient, the probe and the surgical instruments. As a consequence, the preoperative MRI and the post-resection ioUS of a given subject are not initially in arbitrary poses with respect to each other: they \textbf{share a world-frame alignment} inherited from the neuronavigator. Resampling either volume onto the cube grid (Section~\ref{sec:cube}) using its NIfTI affine preserves this alignment.

The world-frame alignment, however, is \emph{not} a registration: it captures only the rigid component of the patient pose between MRI and ioUS acquisitions, and even that is approximate because the patient is reseated between scanner suite and operating room and because the ioUS probe pose is recovered by neuronavigation tracking with millimetre-level uncertainty. Most importantly, it cannot capture \textbf{brain shift}: the deformation of the soft tissue between the preoperative MRI (acquired typically the day before surgery, with the patient supine and the head in a static pose) and the intraoperative ioUS (acquired after craniotomy and partial resection, with cerebrospinal fluid drainage, gravity-induced sag and active tissue removal). The residual landmark $\TRE$ in the cube under the identity transformation is the operational definition of the \emph{initial} configuration that registration must improve upon, and is referred to throughout the remainder of this manuscript as the \textbf{Initial} condition (Section~\ref{sec:reg-compare}).

\subsubsection{Rigid + affine initialisation: NiftyReg block-matching}\label{sec:rigid}

A coarse rigid + 12-parameter affine alignment is computed on top of the world-frame alignment with \textbf{NiftyReg \reg{reg\_aladin}} \citep{ourselin_reconstructing_2001,modat_global_2014}, a random-sampling 3D block-matching solver. The fixed image is the post-resection ioUS in the cube; the moving image is the FOV-pre-masked preoperative MRI in the cube (Section~\ref{sec:cube}). The resulting matrix $\mathbf{A}_{rig}$ is shared by every downstream formulation. The FOV pre-masking of the moving image is required for the block-matching consensus to converge to the surgical region rather than to extra-cranial structures; without it, \reg{reg\_aladin} produced spurious $25^\circ$-rotation solutions in a subset of subjects on this cohort.

\subsubsection{Classical nonrigid registration: NiftyReg \reg{reg\_f3d} with MIND-SSC}\label{sec:reg-classical}

The deformable stage is a free-form B-spline deformation field implemented in \textbf{NiftyReg \reg{reg\_f3d}} \citep{modat_fast_2010} with a control-point spacing of 5~mm. Let $F$ and $M$ denote the fixed (ioUS) and moving (rigidly aligned MRI) volumes, and let $\mathbf{T}(\mathbf{r}) = \mathbf{r} + \mathbf{u}(\mathbf{r})$ be the cubic-B-spline transformation parameterised by a regular grid of control points $\{\boldsymbol{\phi}_k\}$. The solver minimises
\begin{equation}\label{eq:f3d-obj}
\mathcal{J}(\mathbf{T}) \;=\; \mathcal{D}_{\text{MIND}}\!\big(F,\, M \circ \mathbf{T}\big) + \lambda_{BE}\, \operatorname{BE}(\mathbf{T}),
\end{equation}
with the MIND-SSC similarity term \citep{heinrich_mind_2013}
\begin{equation}\label{eq:mindssc-sim}
\mathcal{D}_{\text{MIND}}(F, M\circ\mathbf{T}) \;=\; \sum_{\mathbf{r} \in \Omega_{\text{FOV}}}\, \sum_{\mathbf{p} \in \mathcal{N}} \big|\, \text{MIND}_F(\mathbf{r}, \mathbf{p}) - \text{MIND}_{M \circ \mathbf{T}}(\mathbf{r}, \mathbf{p}) \,\big|,
\end{equation}
and the bending-energy regulariser
\begin{equation}\label{eq:be}
\operatorname{BE}(\mathbf{T}) \;=\; \frac{1}{|\Omega|}\, \int_{\Omega}\, \sum_{i \leq j} \left( \frac{\partial^2 \mathbf{T}(\mathbf{r})}{\partial r_i\, \partial r_j} \right)^{\!2}\, d\mathbf{r}.
\end{equation}
The MIND-SSC descriptor at voxel $\mathbf{r}$ encodes the local patch self-similarity profile
\begin{equation}\label{eq:mind}
\text{MIND}(\mathbf{r}, \mathbf{p}) \;=\; \frac{1}{Z(\mathbf{r})}\, \exp\!\left( -\frac{D_p(\mathbf{r},\, \mathbf{r}+\mathbf{p})}{V(\mathbf{r})} \right), \quad \mathbf{p} \in \mathcal{N},
\end{equation}
where $\mathcal{N}$ is a fixed six to twelve element neighbourhood, $D_p$ is the patch sum-of-squared-differences with a self-similarity-context kernel, $V(\mathbf{r})$ is the local variance of $D_p$ over $\mathcal{N}$ and $Z(\mathbf{r})$ normalises $\max_{\mathbf{p} \in \mathcal{N}} \text{MIND}(\mathbf{r}, \mathbf{p}) = 1$. The descriptor is by construction invariant to modality-specific monotonic intensity transformations.

The solver is restricted to the FOV cone interior by passing $\mathbb{1}_{\Omega_{\text{FOV}}}$ as the \reg{reg\_f3d -rmask} argument on the fixed image; we use a control-point spacing of 5~mm and $\lambda_{BE} = 5 \times 10^{-3}$. The output is a dense per-voxel displacement field $\mathbf{u}_{NR}(\mathbf{r})$ expressed in millimetres with sign convention $-1$ (apply as $\mathbf{r} - \mathbf{u}_{NR}(\mathbf{r})$).

\subsubsection{Synthesis-anchored learning-based registration: SynthMorph}\label{sec:reg-sm}

The learning-based displacement-field network is \textbf{SynthMorph} \citep{hoffmann_synthmorph_2022,hoffmann_anatomy_2024}, pre-trained on a large multi-modal MRI corpus with synthetic intensity augmentation. SynthMorph takes a (fixed, moving) pair of intensity volumes and emits a displacement in voxel units with sign convention $+1$ (apply as $\mathbf{r} + \mathbf{u}(\mathbf{r})$). Applied directly to the raw (ioUS, preoperative MRI) pair, the network produces small-magnitude displacements because the ioUS and MRI intensity distributions are far apart in its training prior. We therefore substitute the synthetic MRI for the ioUS as fixed image. SynthMorph registers the synthetic MRI against the rigidly aligned preoperative MRI; this \textbf{synthesis anchoring} collapses the inter-modality problem into an MRI-versus-MRI problem inside the FOV cone, which the network handles cleanly.

Two operating modes are evaluated: a single-scale forward pass at native cube resolution, and a coarse-to-fine schedule that processes the volumes at $1/4$, $1/2$ and $1\times$ resolution and either composes the displacements sequentially or fuses them in parallel with fixed weights $(0.55, 0.30, 0.15)$ favouring the finest scale. Five fixed-image weighting sub-variants are also evaluated as ablations on the synthesis-only configuration: uniform weighting; cavity voxels zeroed; only parenchyma voxels retained (cavity and tumour both zeroed); tumour voxels weighted $\times 2$; and an exponential spatial decay (10~mm) from the cavity boundary with tumour boost $\times 1.5$. The tumour and resection-cavity segmentations used to define these sub-variants were generated semi-automatically inside ITK-SNAP with the nnInteractive plug-in and reviewed by a board-certified neurosurgeon, following the protocol of the companion benchmark \citep{estebansinovas_inprep}. These segmentations are used only for the retrospective evaluation of the SynthMorph stage and for the landmark stratification of Section~\ref{sec:tre}; the deployed pipeline uses uniform weighting and therefore does not require any segmentation at inference time. The headline configuration uses uniform weighting in single-scale mode.

\subsubsection{Proposed pipeline: NiftyReg + SynthMorph}\label{sec:reg-proposed}

The proposed configuration is the two-stage chain \reg{reg\_aladin}~(rigid+affine)~+~\reg{reg\_f3d}~(MIND-SSC)~+~SynthMorph~(synth-anchored, single-scale, uniform weighting). Concretely: NiftyReg \reg{reg\_aladin} estimates the rigid+affine matrix between the post-resection ioUS (fixed) and the FOV-pre-masked preoperative MRI (moving); NiftyReg \reg{reg\_f3d} estimates the B-spline displacement field on top of the rigidly aligned pair with the MIND-SSC similarity term; SynthMorph then refines the result by registering the synthetic MRI (fixed) against the \reg{reg\_f3d}-warped preoperative MRI (moving), producing the final residual voxel displacement. The composed warp consists of the NiftyReg millimetre-displacement followed by the SynthMorph voxel-displacement.

This formulation is motivated by two complementary observations from the development phase. First, the NiftyReg stage with MIND-SSC achieves the lowest mean $\TRE$ in our cohort, but the pipeline requires more than a millimetre figure on landmarks: it needs a warped preoperative MRI that can be composed with the synthesis output on the shared cube grid, plus a forward transformation that can be inverted reliably to propagate the cube-frame composition back to the patient's preoperative imaging space. The NiftyReg stage alone provides the warped MRI but not the synthesis-driven anchoring that lets the learning residual remain consistent with the synthetic appearance inside the FOV cone. Second, SynthMorph applied alone after a rigid initialisation produces small-magnitude displacements that do not fully recover the bulk brain-shift deformation; when applied on top of the NiftyReg-warped MRI, however, it acts as a fine residual correction in an MRI-versus-MRI setting and produces a smooth, synthesis-consistent refinement.

The role of this two-stage chain in the present manuscript is therefore not to improve $\TRE$ over the NiftyReg baseline but to provide an MRI-versus-MRI-anchored warped preoperative MRI that can serve as the canvas of the siMRI composition. The $\TRE$-versus-NiftyReg comparison is reported in Section~\ref{sec:results-reg} with the explicit statement that the two formulations are statistically indistinguishable on the landmark endpoint (paired Wilcoxon $p = 0.76$ on the present 14-subject cohort) and that both improve significantly over the Initial world-frame alignment ($p \le 0.04$).

\subsubsection{Comparison configurations}\label{sec:reg-compare}

Two further configurations are evaluated as comparison baselines.

\textbf{Initial} (no registration). The identity transformation $\Phi_0(\mathbf{y}) = \mathbf{A}_{cube}^{-1}\, \mathbf{A}_{\text{MR}}\, \mathbf{y}$ applied to a preoperative MRI voxel. The cube-resampled preoperative MRI and ioUS-post share the world-frame alignment inherited from the neuronavigator (Section~\ref{sec:starting}); the Initial $\TRE$ measures the residual landmark error under this starting condition and is the reference against which every registration formulation is compared. Inclusion of Initial in the comparison answers the operationally meaningful question of whether registration improves on what the neuronavigator already gives the surgeon, and quantifies the magnitude of the brain-shift component that no rigid alignment can recover.

\textbf{Synth-as-fixed NiftyReg control}: NiftyReg \reg{reg\_aladin}~+~\reg{reg\_f3d} with the synthetic MRI substituted as fixed image throughout, used as a control to test whether MIND-SSC benefits from the synthesis bridge.

\subsubsection{Coordinate-system contract}\label{sec:coord}

Mixing classical and learning-based displacement fields is error-prone because three independent conventions intersect: the NiftyReg \reg{reg\_aladin} affine matrix $\mathbf{A}_{rig}$ (estimated fixed-to-moving and inverted for forward propagation), the NiftyReg \reg{reg\_f3d} displacement field $\mathbf{u}_{NR}$ in millimetres with sign convention $-1$, and the SynthMorph displacement field $\mathbf{u}_{SM}$ in voxels with sign convention $+1$. Let $\mathbf{A}_{\text{MR}}$ be the voxel-to-world affine of the preoperative MRI, $\mathbf{A}_{cube}$ the voxel-to-world affine of the cube, and $\mathbf{R}_{cube}$ the rotation part of $\mathbf{A}_{cube}$ (so that $\mathbf{R}_{cube}^{-1}$ converts millimetre displacements to voxel displacements). The forward propagation of a preoperative MRI voxel $\mathbf{y}$ through the full pipeline is the composition $\Phi = \Phi_{SM} \circ \Phi_{NR} \circ \Phi_{rig}$, defined stagewise:
\begin{align}
\mathbf{c}_0 &\;=\; \mathbf{A}_{cube}^{-1}\, \mathbf{A}_{rig}^{-1}\, \mathbf{A}_{\text{MR}}\, \mathbf{y}, \label{eq:chain-aff}\\
\mathbf{c}_1 &\;=\; \mathbf{c}_0 - \mathbf{R}_{cube}^{-1}\, \mathbf{u}_{NR}(\mathbf{c}_0), \label{eq:chain-nr}\\
\Phi(\mathbf{y}) \;=\; \mathbf{c}_2 &\;=\; \mathbf{c}_1 - \mathbf{u}_{SM}(\mathbf{c}_1). \label{eq:chain-sm}
\end{align}
The two displacement signs are reconciled at the composition step: $\mathbf{u}_{NR}$ is converted from millimetres to voxels by $\mathbf{R}_{cube}^{-1}$ and applied as a subtraction; $\mathbf{u}_{SM}$, although emitted in the convention $\mathbf{r} + \mathbf{u}$, is applied here as a subtraction because forward sampling of the composed transformation propagates preoperative voxels to the cube grid.

\subsection{Synthetic intraoperative MRI composition}\label{sec:pseudo-mr}

The siMRI is the deliverable of the pipeline: a whole-brain volume that looks like an intraoperative MRI but is built entirely from the patient's preoperative MRI and a single intraoperative ioUS acquisition, together with the trained synthesis and registration models of Sections~\ref{sec:synth} and \ref{sec:reg}. Two variants are produced per subject, both at cube resolution and projected back to the preoperative imaging space.

\subsubsection{Cube-frame composition}\label{sec:pseudo-cube}

Let $\hat{M}(\mathbf{r}) = M\!\big(\Phi(\mathbf{r})\big)$ denote the preoperative MRI warped to the cube by the composed transformation of Section~\ref{sec:coord}, and $S(\mathbf{r})$ the synthesis output sampled into the cube as in Section~\ref{sec:synth-inference}. The \textbf{synthesis variant} $P_{\text{synth}}$ is built in two steps, each operating only on the FOV cone $\Omega_{\text{FOV}}$ and requiring no tumour or cavity segmentation.

\textbf{(i) Histogram matching.} The synthesis output is linearly rescaled into the intensity window of $\hat{M}$ over the FOV cone:
\begin{equation}\label{eq:hist-match}
S'(\mathbf{r}) \;=\; \frac{\hat{M}^{(p_{99})} - \hat{M}^{(p_1)}}{S^{(p_{99})} - S^{(p_1)}}\,\big(S(\mathbf{r}) - S^{(p_1)}\big) + \hat{M}^{(p_1)},
\end{equation}
with the percentiles $S^{(\cdot)}, \hat{M}^{(\cdot)}$ computed on $\Omega_{\text{FOV}}$.

\textbf{(ii) FOV feathering.} Inside the FOV cone the deliverable is the histogram-matched synthesis output itself: the synthesis content is not diluted with the warped preoperative MRI, so that the in-cone image reflects the post-resection anatomy recovered from the ioUS rather than a mixture with the preoperative tissue. Outside the cone the synthesis output is meaningless; the cone boundary is feathered by a Gaussian-softened indicator and exterior voxels fall back to $\hat{M}$:
\begin{equation}\label{eq:fov-feather}
P_{\text{synth}}(\mathbf{r}) \;=\; w(\mathbf{r})\, S'(\mathbf{r}) + (1 - w(\mathbf{r}))\, \hat{M}(\mathbf{r}), \quad
w(\mathbf{r}) \;=\; G_{\sigma=5\,\text{mm}} * \mathbb{1}_{\Omega_{\text{FOV}}}(\mathbf{r}),
\end{equation}
where $G_{\sigma} *$ denotes Gaussian convolution with standard deviation $\sigma$ and $\mathbb{1}_{\Omega_{\text{FOV}}}$ is the indicator function of the FOV cone, so that $w$ is $1$ deep inside the cone, decays smoothly to $0$ across the boundary, and depends on no segmentation label. The result is saved in the cube affine, in $[0, 1]$ intensity range.

Because the in-cone deliverable equals the synthesis output, its fidelity to the intraoperative anatomy is exactly the synthesis fidelity quantified against the intraoperative T2 in Section~\ref{sec:results-synth-quality} (Table~\ref{tab:synth-quality}); the warped preoperative MRI enters the siMRI only as extra-FOV context and through the back-projection transformation of Section~\ref{sec:wholebrain}, not as an in-cone blend.

The \textbf{ultrasound-insert variant} $P_{\text{USinsert}}$ shares the cube grid with $P_{\text{synth}}$ (it inherits the same proposed-pipeline registration of preoperative MRI to ioUS-post). Inside the FOV cone the cube is filled by a linear histogram match of the ioUS to the percentile window of the warped preoperative MRI $\hat M$ so that the inserted speckle is visually compatible with the surrounding MRI brightness; outside the cone the cube falls back to $\hat M$. The FOV-feathering step of Equation~\eqref{eq:fov-feather} is unchanged.

\subsubsection{Whole-brain back-projection}\label{sec:wholebrain}

The whole-brain siMRI on the patient's preoperative imaging space is built by forward-mapping the cube-frame $P_{\text{synth}}$ (which inside the FOV cone is the histogram-matched synthesis content and outside the FOV cone is the preoperative MRI itself) through the composition $\Phi$ of Section~\ref{sec:coord}. For each preoperative MRI voxel $\mathbf{y}$, the forward-mapped position $\Phi(\mathbf{y})$ in the cube is computed and the cube-frame siMRI is sampled there; voxels whose forward image falls outside the cube retain the original preoperative MRI intensity. The cone-projection mask is feathered in the preoperative imaging space to avoid an intensity step at the FOV edge:
\begin{equation}\label{eq:wholebrain}
P_{\text{whole}}(\mathbf{y}) \;=\; w_{\text{whole}}(\mathbf{y})\, P_{\text{synth}}\!\big(\Phi(\mathbf{y})\big) + (1 - w_{\text{whole}}(\mathbf{y}))\, M(\mathbf{y}),
\end{equation}
\begin{equation}\label{eq:wholebrain-w}
w_{\text{whole}}(\mathbf{y}) \;=\; G_{\sigma=5\,\text{mm}} * \mathbb{1}_{\{\Phi(\mathbf{y}) \in \Omega_{\text{FOV}}\}}(\mathbf{y}).
\end{equation}

The result is a whole-brain volume whose intra-FOV content reflects the intraoperative anatomy and whose extra-FOV content is identical to the patient's preoperative MRI. It is saved with the original preoperative MRI affine and constitutes the volume that downstream surgical-navigation systems can load directly.

\subsection{Synthesis-confidence map}\label{sec:confidence}

Because the in-FOV deliverable is the synthesis output rather than a blend anchored to the patient's real anatomy (Section~\ref{sec:pseudo-cube}), the pipeline additionally produces a per-voxel confidence map that signals, at each location inside the FOV cone, how reliable the synthesis prediction is. The map is an interpretable deployment aid, not a calibrated probability of error; its validation against synthesis error is reported in Section~\ref{sec:results-confidence}.

\subsubsection{Test-time-augmentation confidence}\label{sec:confidence-tta}

The confidence map is derived from the disagreement of ResViT-2.5D under input transformations to which it should be equivariant. ResViT-2.5D is run $N = 8$ times on the same subject under the in-plane dihedral symmetry group $D_4$ (the four $90^\circ$ rotations and their horizontal-flip composites). Each element is an exact in-plane symmetry that requires no interpolation, so its inverse on the synthesis output is also exact and the per-voxel correspondence across passes is preserved; the axial slice-context direction of the 2.5D inference is preserved by every $D_4$ element. Let $\sigma_{\text{TTA}}(\mathbf{r})$ be the standard deviation across the eight de-augmented predictions at voxel $\mathbf{r}$. The confidence is the robust-normalised complement of this disagreement,
\begin{equation}\label{eq:confidence}
c(\mathbf{r}) \;=\; G_{\sigma=2\,\text{mm}} * \Big(\, 1 - \operatorname{clip}\!\Big(\tfrac{\sigma_{\text{TTA}}(\mathbf{r}) - q_{10}}{\max(q_{90} - q_{10},\, \varepsilon)},\, 0,\, 1\Big) \Big),
\end{equation}
where $q_{10}$ and $q_{90}$ are the $10$th and $90$th percentiles of $\sigma_{\text{TTA}}$ over the FOV cone; the result is Gaussian-smoothed, clipped to $[0, 1]$ and masked to the cone. Voxels above the $90$th percentile of disagreement saturate to zero confidence. Using the dispersion of equivariant test-time augmentations as a model-uncertainty proxy follows Wang et al. \citep{wang_tta_2019}: where ResViT-2.5D has not internalised the $D_4$ symmetry, its output fluctuates across augmentations and the model is signalling uncertainty.

\subsubsection{Disclosure flag}\label{sec:confidence-flags}

A further per-voxel signal is produced to disclose a specific, clinically meaningful failure mode that the TTA confidence does not separate: an \emph{input-quality flag} $s_{\text{input}}$, a sigmoid of the local signal-to-noise ratio of the ioUS (computed from $5$-voxel local mean and variance) centred on the lower quartile of the in-FOV distribution. It scores near $1$ on clean parenchymal speckle and near $0$ on acoustic shadows and the FOV-cone edge where the ioUS carries no signal. As shown in Section~\ref{sec:results-confidence}, this flag does not predict the magnitude of synthesis pixel error, and it is therefore kept as a separate disclosure channel rather than aggregated into the confidence map.

\subsubsection{Operating-room visualisation}\label{sec:confidence-orview}

For intraoperative display we produce a volume that bakes the siMRI grayscale (at $60\%$ of the dynamic range to leave headroom) under a single red overlay proportional to $(1 - c)^{1.5}$ (TTA uncertainty), so that low-confidence regions are highlighted directly on the anatomy. The volume is produced both in the cube frame and forward-mapped to the preoperative MRI frame for direct loading on the whole-brain siMRI. The overlay carries only the error-validated TTA confidence; the input-quality flag $s_{\text{input}}$ is retained solely as a quantitative diagnostic in the offline analysis (Section~\ref{sec:results-confidence}) and is not rendered in any visualisation.

\subsection{Evaluation protocol}\label{sec:eval}

Three independent endpoints are evaluated: registration accuracy by landmark $\TRE$; image-similarity quality of synthesis and siMRI inside the FOV cone; and deformation-field geometric plausibility on the composed transformation.

\subsubsection{Landmark target registration error}\label{sec:tre}

For each pair $(\mathbf{p}_i^{\text{MR}},\, \mathbf{p}_i^{\text{ioUS}})$ of a subject $s$ with $N_s$ valid pairs (Section~\ref{sec:splits}), the MRI fiducial is forward-propagated through $\Phi_s$ (Section~\ref{sec:coord}) to the cube and the per-pair $\TRE$ is computed in millimetres. For the \textbf{Initial} condition (Section~\ref{sec:reg-compare}), $\Phi_s$ reduces to the identity world-frame mapping $\Phi_0(\mathbf{y}) = \mathbf{A}_{cube}^{-1}\, \mathbf{A}_{\text{MR}}\, \mathbf{y}$; for every other formulation $\Phi_s$ includes the corresponding NiftyReg and/or SynthMorph stages. The per-pair $\TRE$ is
\begin{equation}\label{eq:tre-pair}
\TRE_{s,i} \;=\; \Delta_{cube}\, \big\| \Phi_s(\mathbf{p}_i^{\text{MR}}) - \mathbf{p}_i^{\text{ioUS}} \big\|_2,
\end{equation}
with $\Delta_{cube} = 0.5$~mm the cube voxel size. We report per subject the mean and the robust worst-30\% endpoint adopted by the Learn2Reg community:
\begin{equation}\label{eq:tre-subj}
\TRE(s) \;=\; \frac{1}{N_s}\, \sum_{i=1}^{N_s} \TRE_{s,i}, \qquad
\TRE_{30}(s) \;=\; \mathrm{percentile}_{70}\!\left( \{\TRE_{s,i}\}_{i=1}^{N_s} \right).
\end{equation}
We further stratify each pair by its Euclidean distance to the cavity-or-tumour boundary, $d_i = \min_{\mathbf{q} \in \partial(\Omega_T \cup \Omega_C)} \| \mathbf{p}_i^{\text{ioUS}} - \mathbf{q} \|_2$, where $\Omega_T$ and $\Omega_C$ denote the tumour and resection-cavity masks of the ReMIND release used only for this evaluation-time stratification, and define
\begin{equation}\label{eq:tre-stratified}
\TRE\text{-near}(s) \;=\; \operatorname*{mean}_{i : d_i < 15\,\text{mm}} \TRE_{s,i}, \quad
\TRE\text{-far}(s) \;=\; \operatorname*{mean}_{i : d_i \geq 15\,\text{mm}} \TRE_{s,i}.
\end{equation}
$\TRE$-near and $\TRE$-far isolate the peri-cavity region from regions in which the preoperative MRI remains essentially valid.

\subsubsection{Image-similarity metrics}\label{sec:img-sim}

Image-similarity metrics are computed inside the FOV cone eroded by one voxel, denoted $\Omega_{\text{FOV}}^{\ominus}$. Both volumes are mapped to $[0, 1]$ by percentile anchoring at $(p_1, p_{99})$ of the reference $\mathbf{y}$:
\begin{equation}\label{eq:perc-anchor}
\tilde{\mathbf{y}}(\mathbf{r}) \;=\; \operatorname{clip}\!\left( \frac{\mathbf{y}(\mathbf{r}) - p_1}{p_{99} - p_1};\, 0,\, 1 \right), \quad \tilde{\hat{\mathbf{y}}}\text{ is rescaled identically.}
\end{equation}
The peak signal-to-noise ratio ($\PSNR$) is computed globally on $\Omega_{\text{FOV}}^{\ominus}$ from the mean-squared-error:
\begin{equation}\label{eq:psnr}
\mathrm{MSE} \;=\; \frac{1}{|\Omega_{\text{FOV}}^{\ominus}|}\, \sum_{\mathbf{r} \in \Omega_{\text{FOV}}^{\ominus}}\, \big(\tilde{\hat{\mathbf{y}}}(\mathbf{r}) - \tilde{\mathbf{y}}(\mathbf{r})\big)^2, \quad
\PSNR \;=\; 10\, \log_{10}\!\left( \frac{1}{\mathrm{MSE}} \right).
\end{equation}
The mean absolute error ($\MAE$) is the $L_1$ analogue,
\begin{equation}\label{eq:mae}
\MAE \;=\; \frac{1}{|\Omega_{\text{FOV}}^{\ominus}|}\, \sum_{\mathbf{r} \in \Omega_{\text{FOV}}^{\ominus}}\, \big| \tilde{\hat{\mathbf{y}}}(\mathbf{r}) - \tilde{\mathbf{y}}(\mathbf{r}) \big|.
\end{equation}
The normalised cross-correlation ($\NCC$) is averaged over axial slices $z \in Z$ inside the mask, with $M_z = \Omega_{\text{FOV}}^{\ominus} \cap \{z\}$, $v = \tilde{\mathbf{y}}$, $\hat{v} = \tilde{\hat{\mathbf{y}}}$, and $\bar v_z, \bar{\hat v}_z$ their slice-wise means restricted to $M_z$:
\begin{equation}\label{eq:ncc}
\NCC \;=\; \frac{1}{|Z|}\, \sum_{z \in Z}\, \frac{ \sum_{\mathbf{r} \in M_z}\,(\hat{v}(\mathbf{r}) - \bar{\hat{v}}_z)\, (v(\mathbf{r}) - \bar{v}_z) }{ \sqrt{ \sum_{\mathbf{r} \in M_z}(\hat{v}(\mathbf{r}) - \bar{\hat{v}}_z)^2\, \sum_{\mathbf{r} \in M_z}(v(\mathbf{r}) - \bar{v}_z)^2 } }.
\end{equation}
The structural similarity index \citep[$\SSIM$;][]{wang_ssim_2004} is computed slice-wise along the axial axis with $\text{data\_range} = 1.0$ on a sliding $7 \times 7$ window:
\begin{equation}\label{eq:ssim}
\SSIM(\hat{v}, v) \;=\; \frac{(2\,\mu_{\hat v}\,\mu_v + c_1)\,(2\,\sigma_{\hat v v} + c_2)}{(\mu_{\hat v}^2 + \mu_v^2 + c_1)\,(\sigma_{\hat v}^2 + \sigma_v^2 + c_2)},
\end{equation}
with $c_1 = (0.01\,L)^2$, $c_2 = (0.03\,L)^2$, $L = 1$ the dynamic range, $\mu_\cdot, \sigma_\cdot, \sigma_{\hat v v}$ the local means, variances and covariance over the window. The metric is averaged over those slices for which $|M_z|/|\{z\}| > 1\%$. The learned perceptual image patch similarity \citep[$\LPIPS$;][]{zhang_lpips_2018} is the calibrated $\ell_2$ distance between AlexNet feature maps,
\begin{equation}\label{eq:lpips}
\LPIPS(\hat v, v) \;=\; \sum_{l \in \mathcal{L}}\, \frac{1}{H_l W_l}\, \sum_{h, w}\, \big\| \mathbf{w}_l \odot \big(\phi_l(\hat v)_{h, w} - \phi_l(v)_{h, w}\big) \big\|_2^2,
\end{equation}
with $\phi_l$ the AlexNet feature map at layer $l \in \mathcal{L}$, $\mathbf{w}_l$ the per-channel calibration weights of Zhang et al. \citep{zhang_lpips_2018} and the input slices replicated to three channels and rescaled to $[-1, 1]$.

The metrics are computed (a) between the synthesis output and the intraoperative T2 resampled to the cube for the synthesis-quality analysis of Section~\ref{sec:results-synth-quality} on the 14-subject post-resection cohort; (b) between the synthesis output and the NiftyReg-warped preoperative T2 for the spacing-robustness analysis of Section~\ref{sec:results-spacing-shift}, where the warped preoperative T2 is the reference whose acquisition spacing does not vary across the perturbation cells and therefore isolates the synthesis-side variation; (c) between the synthesis-variant siMRI (Psynth) and the intraoperative T2 inside the FOV cone, which under the deployed composition equals the in-cone synthesis output and is therefore quantified directly in Section 4.1.1 (Table 2); the qualitative inspection of both siMRI variants is reported in Section~\ref{sec:results-pseudo}.

\subsubsection{Deformation-field geometric plausibility}\label{sec:plausibility}

For each candidate registration formulation we evaluate the \textbf{composed transformation} $\Phi$ that is actually applied at deployment. The Jacobian determinant of $\Phi$ at a voxel $\mathbf{r}$ is
\begin{equation}\label{eq:jacobian}
J_{\Phi}(\mathbf{r}) \;=\; \det \nabla \Phi(\mathbf{r}),
\end{equation}
approximated by central finite differences with reflective padding in single precision. The \textbf{folding fraction} counts the voxels inside the FOV for which $\Phi$ is locally non-injective,
\begin{equation}\label{eq:folding}
\text{folding}\% \;=\; 100\, \cdot\, \frac{ \big| \{\mathbf{r} \in \Omega_{\text{FOV}} :\, J_{\Phi}(\mathbf{r}) \leq 0\} \big| }{ \big|\Omega_{\text{FOV}}\big| },
\end{equation}
and the \textbf{standard deviation of $\log |J|$} is computed over the diffeomorphic part of the field,
\begin{equation}\label{eq:sdlogj}
\mathrm{SDLogJ} \;=\; \operatorname{std}\!\big(\, \log |J_{\Phi}(\mathbf{r})| \,:\, \mathbf{r} \in \Omega_{\text{FOV}},\; J_{\Phi}(\mathbf{r}) > 0 \,\big).
\end{equation}
We explicitly distinguish between the Jacobian metrics of the SynthMorph stage in isolation (reported once in Section~\ref{sec:results-plausibility} as a diagnostic of the learning residual) and those of the composed transformation (the operationally relevant quantity, reported in the main results table).

\subsubsection{Statistical analysis}\label{sec:stats}

All paired comparisons across the 14 test subjects use the two-sided Wilcoxon signed-rank test on per-subject values; no parametric test is used. For a pair of methods $A, B$ and a metric $m$ we report the median paired difference $\Delta = \operatorname{median}_s\!\big(m_A(s) - m_B(s)\big)$ and the two-sided $p$-value. No multiple-comparison correction is applied within a single hypothesis; cases in which a finding is borderline are flagged explicitly. For the spacing-robustness analysis of Section~\ref{sec:results-spacing-shift}, given a synthesis backbone, a metric $m$ and a subject $s$ evaluated at a set of acquisition spacings $\{\mathbf{h}_k\}$ with $\mathbf{h}_0$ the cube native, we define the per-spacing log-distance
\begin{equation}\label{eq:log-dist}
d_k \;=\; \big\| \log \mathbf{h}_k - \log \mathbf{h}_0 \big\|_2,
\end{equation}
and fit a robust linear regression of $m_s$ on $d_k$ with a Huber loss $\rho_H$,
\begin{equation}\label{eq:huber}
(\hat{a}_s,\, \hat{b}_s) \;=\; \arg\min_{a,b}\, \sum_{k}\, \rho_H\!\big(\, m_s(\mathbf{h}_k) - (a + b\, d_k) \,\big),
\end{equation}
so that $|\hat{b}_s|$ is the per-subject sensitivity index to spacing shift. The sensitivity indices are then compared pairwise across synthesis backbones by paired Wilcoxon.

\subsection{Software and compute environment}\label{sec:software}

The pipeline orchestration, intensity preprocessing and siMRI composition are implemented in Python~3.10. The synthesis network was trained in PyTorch~2.5.1 (CUDA~12.1, cuDNN~9) on a single NVIDIA RTX~3090 GPU (24~GB VRAM). The classical registration backend is NiftyReg~1.5.58 (\reg{reg\_aladin} and \reg{reg\_f3d} with MIND-SSC similarity); the learning-based displacement-field stage is SynthMorph (affine model v2, deformable model v3), distributed through a FreeSurfer development build (\texttt{freesurfer-linux-ubuntu22\_x86\_64-dev-20260416}, commit \texttt{18761d3}). Both back-ends run under Windows Subsystem for Linux.

\paragraph{End-to-end wallclock per subject.} We measured per-stage wallclock on the 14 test subjects, on the same RTX~3090 hardware, running the pipeline from raw scanner NIfTIs to whole-brain siMRI. The stages timed are the cross-modality synthesis, the rigid+affine \reg{reg\_aladin} stage, the NiftyReg \reg{reg\_f3d} stage, the SynthMorph deformable stage, and the siMRI composition (both cube and back-projection to the preoperative MRI grid). Table~\ref{tab:wallclock} reports the median, mean$\pm$s.d.\ and range across the cohort.

\begin{table}[t]
\centering
\caption{End-to-end wallclock per subject of the present pipeline on the 14-subject test cohort. Times are wall-clock seconds on a single NVIDIA RTX~3090 (24~GB) with NiftyReg running under Windows Subsystem for Linux. Synth covers crop $\rightarrow$ ResViT-2.5D inference $\rightarrow$ resample-to-cube; siMRI covers cube composition $+$ forward-mapped whole-brain projection $+$ US-insert variant.}
\label{tab:wallclock}
\footnotesize
\begin{tabular}{lccc}
\toprule
Stage & Median (s) & Mean $\pm$ s.d.\ (s) & Range (s) \\
\midrule
Preprocess (cube + FOV mask)         & 11.3 & 11.9 $\pm$ 3.6  & 9 -- 19 \\
Synthesis (ResViT-2.5D + resample)   & 26.4 & 27.1 $\pm$ 6.7  & 19 -- 47 \\
\reg{reg\_aladin} (rigid + affine)   & 39.8 & 39.8 $\pm$ 8.3  & 28 -- 53 \\
\reg{reg\_f3d} (MIND-SSC)            & 87.7 & 96.0 $\pm$ 36.2 & 56 -- 196 \\
SynthMorph (deformable)              & 313.0 & 313.0 $\pm$ 39.6 & 226 -- 368 \\
siMRI composition              & 20.5 & 24.5 $\pm$ 9.6  & 13 -- 50 \\
\midrule
\textbf{Total per subject}           & \textbf{508} & \textbf{513 $\pm$ 33} & \textbf{480 -- 563} \\
\bottomrule
\end{tabular}
\end{table}

%% ============================================================================
\section{Results}\label{sec:results}
%% ============================================================================

We report three sets of experiments aligned with the three stages of the pipeline. Section~\ref{sec:results-spacing} quantifies the cross-modality synthesis quality of the three reference backbones (ResViT-2.5D, ResViT-3D and SynDiff-2.5D) and their robustness to acquisition-spacing shift; together with inference cost, these properties inform the choice of deployed backbone. Section~\ref{sec:results-reg} evaluates the registration formulations of Section~\ref{sec:reg} against the world-frame Initial baseline and against the strongest classical baseline, on landmark $\TRE$ and on the geometric plausibility of the composed transformation. Section~\ref{sec:results-pseudo} reports the synthetic intraoperative MRI deliverable on the present cohort, both qualitatively and through coherence checks that operate on the composed siMRI without requiring an external reference image (composed-warp field plausibility, ioUS-versus-synthesis correlation, ioUS-versus-USinsert correlation).

The cross-modality synthesis-quality benchmark of Section~\ref{sec:results-synth-quality} follows the protocol of the prior synthesis benchmark \citep{estebansinovas_inprep} that selected the deployed checkpoint and operates on the 14-subject primary test cohort. All paired comparisons are two-sided Wilcoxon signed-rank tests on per-subject values; no parametric test is used.

\subsection{Synthesis quality and spacing robustness}\label{sec:results-spacing}

\subsubsection{Image-quality benchmark of the three candidate backbones}\label{sec:results-synth-quality}

The three candidate synthesis backbones were evaluated inside the FOV cone of each test subject, with the intraoperative T2 resampled to the cube as reference (Section~\ref{sec:img-sim}). This is the contrast that the surgical team aims to approximate from ioUS at deployment, so any deviation from it is directly informative for the clinical use case. Table~\ref{tab:synth-quality} reports per-subject means across the 14 test subjects.

\begin{table}[t]
\centering
\caption{Synthesis quality of the three candidate backbones on the 14-subject post-resection test cohort, evaluated against the intraoperative T2 inside the FOV cone (mean $\pm$ standard deviation over per-subject values). Best per metric in bold.}
\label{tab:synth-quality}
\begin{tabular}{lcccc}
\toprule
Backbone & $\SSIM \uparrow$ & $\PSNR$ (dB) $\uparrow$ & $\MAE \downarrow$ & $\LPIPS \downarrow$ \\
\midrule
ResViT-2.5D     & $\mathbf{0.710 \pm 0.047}$ & $\mathbf{14.96 \pm 1.02}$ & $\mathbf{0.121 \pm 0.021}$ & $0.165 \pm 0.037$ \\
ResViT-3D       & $0.695 \pm 0.046$          & $14.36 \pm 1.04$          & $0.132 \pm 0.022$          & $\mathbf{0.160 \pm 0.035}$ \\
SynDiff-2.5D    & $0.707 \pm 0.043$          & $14.42 \pm 0.98$          & $0.129 \pm 0.020$          & $0.166 \pm 0.034$ \\
\bottomrule
\end{tabular}
\end{table}

ResViT-2.5D outperforms ResViT-3D on $\SSIM$, $\PSNR$ and $\MAE$, with paired Wilcoxon signed-rank statistics of $p = 0.009$, $p = 0.005$ and $p = 0.003$ respectively (14 subjects). On $\LPIPS$ the two variants are statistically indistinguishable ($\Delta = +0.005$, $p = 0.22$, ResViT-3D nominally lower). SynDiff-2.5D is competitive with ResViT-2.5D on $\SSIM$ and $\LPIPS$ but trails it on $\PSNR$, $\MAE$ . (Section~\ref{sec:results-spacing-shift}). The relative ordering observed here differs from the one reported in Esteban-Sinovas et al. \citep{estebansinovas_inprep}, in which ResViT-3D yielded the highest peak quality against the warped preoperative T2 of the training distribution. The inversion is driven by the reference: against the intraoperative T2, which carries the post-resection cavity and the brain-shifted anatomy that ResViT-3D was not trained to reproduce, the slice-by-slice inference of ResViT-2.5D matches the target intraoperative geometry more faithfully.

\subsubsection{Robustness to acquisition-spacing shift}\label{sec:results-spacing-shift}
A peak-quality figure at the training spacing is not by itself a deployment criterion. Intraoperative ioUS spacing varies across vendors, probes and acquisition protocols, and a clinically reliable backbone must absorb that variability with minimal quality loss. We probed each backbone at five acquisition spacings spanning the realistic envelope of intraoperative ioUS (Table~\ref{tab:spacings}). For each (subject, backbone, spacing) cell, starting from the \emph{raw} scanner ioUS (the source used here, native $\sim 0.13 \times 0.13 \times 0.5$~mm), the ioUS was resampled to the target spacing, the synthesis network was run on that intermediate grid, the synthetic MRI was resampled back to the cube, and the cube image-similarity metrics of Section~\ref{sec:img-sim} were computed against the NiftyReg-warped preoperative T2 reference (again estimated from the raw inputs). The reference is the same warped preoperative T2 across every perturbation cell so that the only source of metric variation is the synthesis output. The experiment yielded $14 \times 3 \times 5 = 210$ cells.
\begin{table}[t]
\centering
\caption{Acquisition spacings probed in the robustness experiment.}
\label{tab:spacings}
\begin{tabular}{lcl}
\toprule
Tag & Spacing (mm) & Comment \\
\midrule
0.5iso          & $0.5 \times 0.5 \times 0.5$ & cube native \\
$0.66 \times 0.66 \times 1$ & $0.66 \times 0.66 \times 1.0$ & mild anisotropic \\
$0.86 \times 0.86 \times 2$ & $0.86 \times 0.86 \times 2.0$ & dominant training spacing \\
1iso            & $1.0 \times 1.0 \times 1.0$ & research isotropic \\
$1 \times 1 \times 3$ & $1.0 \times 1.0 \times 3.0$ & clinical worst case \\
\bottomrule
\end{tabular}
\end{table}
For each backbone, subject and metric we fitted the robust linear regression of Equation~\eqref{eq:huber} and took $|\hat{b}_s|$ as the per-subject sensitivity index. Table~\ref{tab:slopes} reports the per-backbone mean sensitivity together with the paired Wilcoxon test between ResViT-2.5D and ResViT-3D.
\begin{table}[t]
\centering
\caption{Per-subject mean of the sensitivity index $|\hat{b}_s|$ to acquisition-spacing shift, by metric and backbone, and paired Wilcoxon comparison (14 subjects, two-sided). Lower sensitivity is better; the dominant comparison is ResViT-2.5D vs ResViT-3D.}
\label{tab:slopes}
\begin{tabular}{lcccc}
\toprule
Metric & ResViT-2.5D & ResViT-3D & SynDiff-2.5D & $p$ (2.5D vs 3D) \\
\midrule
$\PSNR$  & 0.585  & \textbf{0.593}  & 0.674  & 0.30 \\
$\SSIM$  & \textbf{0.0012} & 0.0028 & 0.0014 & $\mathbf{9 \times 10^{-3}}$ \\
$\LPIPS$ & 0.0126 & \textbf{0.0038} & 0.0087 & $\mathbf{4 \times 10^{-4}}$ \\
$\NCC$   & \textbf{0.0072} & 0.0240 & 0.0124 & $\mathbf{0.038}$ \\
$\MAE$   & 0.016  & \textbf{0.010}  & 0.027  & $\mathbf{0.044}$ \\
\bottomrule
\end{tabular}
\end{table}
In the present conditions the spacing-robustness margin between ResViT-2.5D and ResViT-3D narrows considerably and is no longer monotone across metrics. A pointwise paired Wilcoxon at each spacing (Table~\ref{tab:pointwise}) confirms this picture: ResViT-3D is significantly worse than ResViT-2.5D in PSNR only at the finest spacing ($0.5$~iso), in SSIM at $0.5$~iso and $0.66 \times 0.66 \times 1$~mm, and is consistently better than ResViT-2.5D in LPIPS at $0.5$~iso. The absolute-quality AUC across the spacing axis is highest for SynDiff-2.5D on PSNR and SSIM, with SynDiff-2.5D significantly outperforming both ResViT variants in three of the five metrics (paired Wilcoxon $p < 10^{-3}$ on PSNR-AUC and SSIM-AUC).

\begin{table}[t]
\centering
\caption{Pointwise paired Wilcoxon between ResViT-2.5D and ResViT-3D at each of the five acquisition spacings (14 test subjects). $\Delta$ is the mean per-subject difference (ResViT-2.5D minus ResViT-3D); on $\PSNR$ a positive $\Delta$ favours ResViT-2.5D, on $\LPIPS$ and $\MAE$ a negative $\Delta$ favours ResViT-2.5D.}
\label{tab:pointwise}
\small
\begin{tabular}{lcccccc}
\toprule
Spacing & $\Delta\PSNR$ (dB) & $p$ & $\Delta\LPIPS$ & $p$ & $\Delta\MAE$ & $p$ \\
\midrule
0.5iso                       & $+1.05$ & $\mathbf{4 \times 10^{-4}}$ & $+0.024$ & $\mathbf{1 \times 10^{-4}}$ & $+0.002$ & 0.95 \\
$0.66 \times 0.66 \times 1$  & $-0.02$ & 0.86 & $-0.001$ & 0.71 & $+0.002$ & 0.95 \\
$0.86 \times 0.86 \times 2$  & $+0.07$ & 0.76 & $+0.001$ & 0.71 & $-0.003$ & 0.71 \\
1iso                         & $-0.50$ & 0.013 & $-0.003$ & 0.08 & $-0.014$ & $\mathbf{2 \times 10^{-3}}$ \\
$1 \times 1 \times 3$        & $-0.10$ & 0.86 & $-0.014$ & 0.013 & $-0.013$ & 0.041 \\
\bottomrule
\end{tabular}
\end{table}

\subsubsection{Backbone selection}\label{sec:backbone-selection}

The selection criterion combines the absolute-quality result of Section~\ref{sec:results-synth-quality}, the spacing-robustness result of Section~\ref{sec:results-spacing-shift}, and operational considerations of inference cost and streaming-friendliness. On absolute quality against the intraoperative T2 reference (Table~\ref{tab:synth-quality}), ResViT-2.5D outperforms ResViT-3D on SSIM, PSNR and MAE. On spacing robustness in the present cohort (Table~\ref{tab:slopes}), the two backbones are no longer separated by a uniform gap: ResViT-2.5D wins on SSIM and NCC, ResViT-3D wins on LPIPS and MAE, and the two tie on PSNR. The downstream landmark $\TRE$ does not distinguish the two backbones (Section~\ref{sec:results-ablation}, $p > 0.5$ on our cohort). The operational considerations break the tie: the 2.5D variant is approximately 6$\times$ faster at inference than ResViT-3D ($\sim 26$~s vs $\sim 2$~min per subject on the cube grid), runs at a 2D parameter budget, and is naturally compatible with a slice-streaming variant suitable for real-time acquisition (Section~\ref{sec:discussion-future}). We therefore deploy ResViT-2.5D in the proposed pipeline; ResViT-3D and SynDiff-2.5D are retained as references in the registration ablations of Section~\ref{sec:results-reg}. SynDiff-2.5D in particular, while showing competitive image-quality numbers on our cohort, is dropped from the deployed pipeline because its iterative diffusion sampling step adds latency without a clear downstream benefit on landmark $\TRE$.

\subsection{Registration evaluation}\label{sec:results-reg}

\subsubsection{Main $\TRE$ comparison}\label{sec:results-tre}

Table~\ref{tab:reg-main} compares the \textbf{proposed two-stage NiftyReg + SynthMorph pipeline (ResViT-2.5D backbone)} against three configurations: (i) the Initial world-frame alignment, (ii) the rigid+affine \reg{reg\_aladin} refinement alone, and (iii) the NiftyReg classical baseline (\reg{reg\_aladin}~+~\reg{reg\_f3d} with MIND-SSC). The synth-as-fixed NiftyReg control of Section~\ref{sec:reg-compare} is reported alongside. The proposed pipeline is also instantiated with the ResViT-3D backbone as a sensitivity check. All $\TRE$ values are per-subject means in millimetres; folding fraction and SDLogJ are computed on the composed transformation that is actually applied at deployment.

\begin{table}[t]
\centering
\caption{Registration accuracy and field plausibility on the 14-subject test cohort (215 landmark pairs). Near / far is the mean per-subject TRE on landmarks within / beyond 15~mm of any cavity-or-tumour voxel. Folding and SDLogJ are computed on the composed transformation. $p$ vs Init and $p$ vs NiftyReg are paired Wilcoxon signed-rank tests against the Initial baseline and the NiftyReg classical baseline, respectively. Mean $\pm$ s.d. over per-subject means.}
\label{tab:reg-main}
\footnotesize
\setlength{\tabcolsep}{3pt}
\begin{tabular}{lccccccc}
\toprule
Method & $\TRE$ (mm) & $\TRE_{30}$ (mm) & near / far (mm) & Folding (\%) & SDLogJ & $p$ vs Init & $p$ vs NiftyReg \\
\midrule
Initial (world-frame)                  & $6.27 \pm 2.37$ & $7.02 \pm 2.38$ & 5.81 / 6.14 & ---   & ---   & ---   & 0.003 \\
\reg{reg\_aladin} (rigid+affine)       & $6.22 \pm 2.35$ & $6.96 \pm 2.37$ & 5.73 / 6.12 & ---   & ---   & 0.17  & 0.004 \\
NiftyReg + MIND-SSC                    & $\mathbf{5.85 \pm 2.26}$ & $\mathbf{6.61 \pm 2.24}$ & \textbf{5.25 / 5.80} & 0.000 & 0.007 & $\mathbf{0.003}$ & --- \\
\textbf{Proposed (NR+SM, ResViT-2.5D)} & $5.86 \pm 2.30$ & $6.57 \pm 2.22$ & 5.39 / 5.80 & 0.000 & 0.080 & $\mathbf{0.042}$ & 0.76 \\
Proposed (NR+SM, ResViT-3D)            & $\mathbf{5.79 \pm 2.28}$ & $6.57 \pm 2.23$ & 5.27 / 5.74 & 0.000 & 0.073 & $\mathbf{0.009}$ & 0.54 \\
Synth-as-fixed NiftyReg                & $6.04 \pm 2.30$ & $6.76 \pm 2.34$ & 5.49 / 5.96 & 0.000 & 0.005 & $\mathbf{9 \times 10^{-4}}$ & 0.30 \\
\bottomrule
\end{tabular}
\end{table}

Three quantitative observations follow.

\textbf{(i) Registration matters on top of the neuronavigator world frame.} The Initial condition, i.e., the NIfTI world-frame alignment that the surgical team already has on the neuronavigator console, yields a per-pair mean $\TRE$ of $6.27$~mm on this cohort. The rigid+affine \reg{reg\_aladin} refinement reduces this by $0.05$~mm only and is not statistically significant ($p = 0.17$); the rigid+affine block-matching alone is not sufficient. The classical nonrigid stage (NiftyReg \reg{reg\_f3d} with MIND-SSC) reduces $\TRE$ to $5.85$~mm, a $0.42$~mm improvement over Initial (paired Wilcoxon $p = 3 \times 10^{-3}$), confirming that a fully deformable formulation is required.

\begin{figure}[!htbp]
  \centering
  \includegraphics[width=\linewidth,height=0.82\textheight,keepaspectratio]{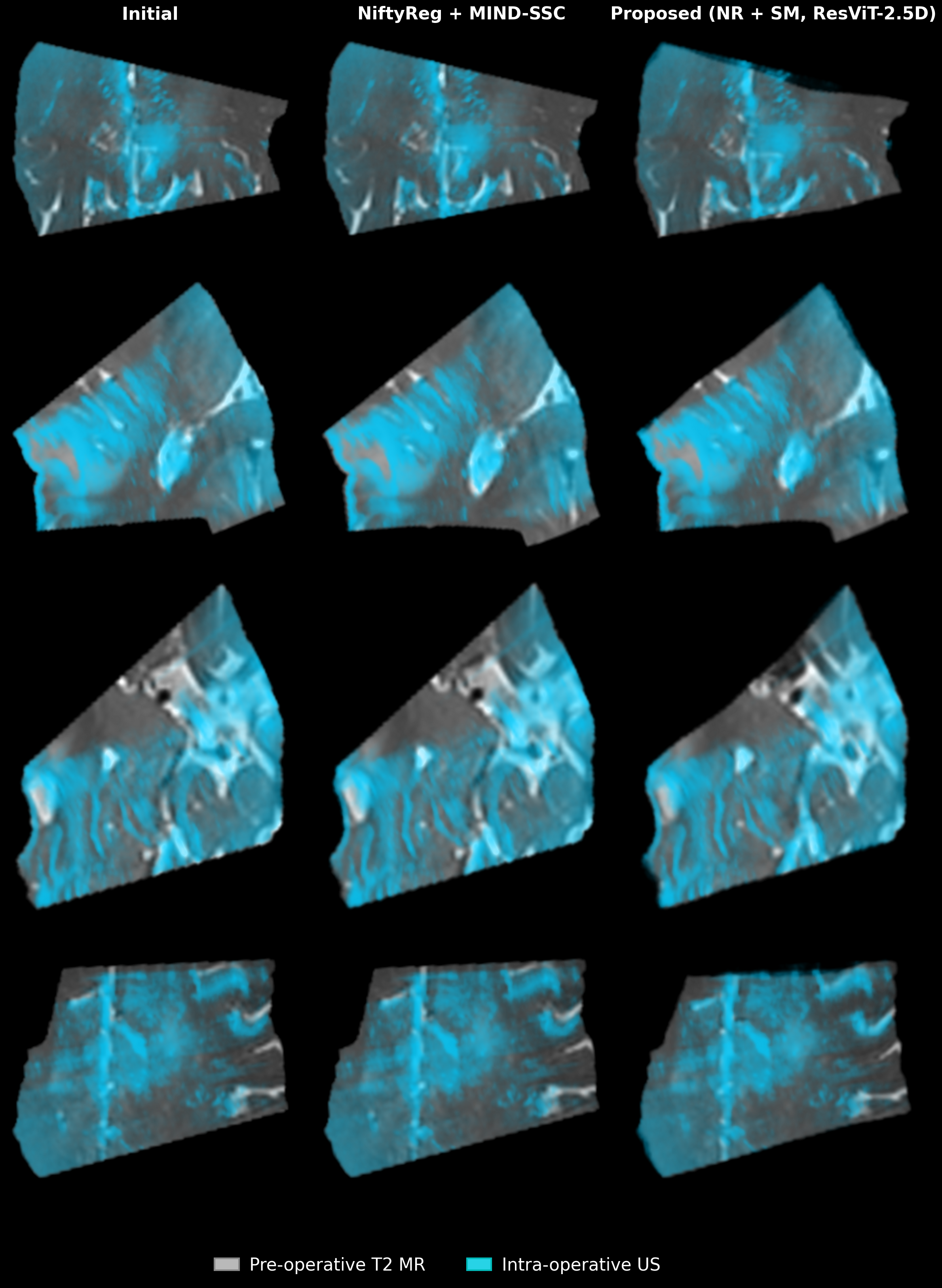}
  \caption{Qualitative registration across four representative subjects
  (rows). Each panel overlays the intraoperative ultrasound (cyan) on the
  preoperative T2-weighted MRI (grey) inside the ultrasound field of view.
  \textbf{Left}: identity transformation on the neuronavigator world frame
  (Initial); \textbf{centre}: NiftyReg classical baseline
  (\reg{reg\_aladin}~+~\reg{reg\_f3d} with MIND-SSC); \textbf{right}: proposed
  two-stage pipeline (NiftyReg + synthesis-anchored SynthMorph, ResViT-2.5D
  backbone). Tighter cyan--grey coincidence indicates better alignment.}
  \label{fig:registration}
\end{figure}

\textbf{(ii) The proposed pipeline matches the classical baseline on $\TRE$ but does not improve upon it.} The two-stage NiftyReg + SynthMorph (ResViT-2.5D) composition yields a per-pair mean $\TRE$ of $5.86$~mm, statistically indistinguishable from the classical NiftyReg baseline (paired Wilcoxon $p = 0.76$, $\Delta = +0.02$~mm). The proposed pipeline beats the Initial condition by $0.41$~mm (paired Wilcoxon $p = 0.04$; Figure~\ref{fig:registration}). Instantiating the same proposed schedule with the ResViT-3D backbone yields a marginally lower $\TRE$ of $5.79$~mm, nominally below the classical baseline (paired Wilcoxon $p = 0.54$, indistinguishable). The role of the synthesis-anchored learning stage in the present pipeline is therefore not to improve $\TRE$ over the classical baseline; it is to produce the warped preoperative MRI on which the siMRI composition of Section~\ref{sec:results-pseudo} is built.

\textbf{(iii) The synth-as-fixed NiftyReg control does not benefit from the modality bridge.} Substituting the synthetic MRI for the ioUS as the fixed image in the classical stage produces $\TRE$ $6.04$~mm versus $5.85$~mm for the ioUS-fixed baseline ($p = 0.30$, classical with ioUS-fixed trending better), consistent with the MIND-SSC similarity term being already modality-invariant.

\subsubsection{Geometric plausibility of the composed transformation}\label{sec:results-plausibility}

Both the classical NiftyReg \reg{reg\_f3d} baseline and the proposed two-stage pipeline produce composed transformations that are fold-free across the 14 test subjects (folding fraction $0.000\%$ on every subject; Table~\ref{tab:reg-main}). SDLogJ differs across methods: the classical baseline yields an extremely smooth composed warp (SDLogJ $0.007$), reflecting that the rigid+affine \reg{reg\_aladin} stage absorbs most of the rigid component and the \reg{reg\_f3d} stage adds only a small residual deformation. The proposed pipeline yields a moderately smooth composed warp (SDLogJ $0.080$ for the ResViT-2.5D variant, $0.073$ for the 3D variant), reflecting the larger non-rigid residual produced by the synthesis-anchored SynthMorph stage on top of the warped preoperative MRI.

\begin{figure}[!htbp]
  \centering
  \includegraphics[width=\linewidth,height=0.82\textheight,keepaspectratio]{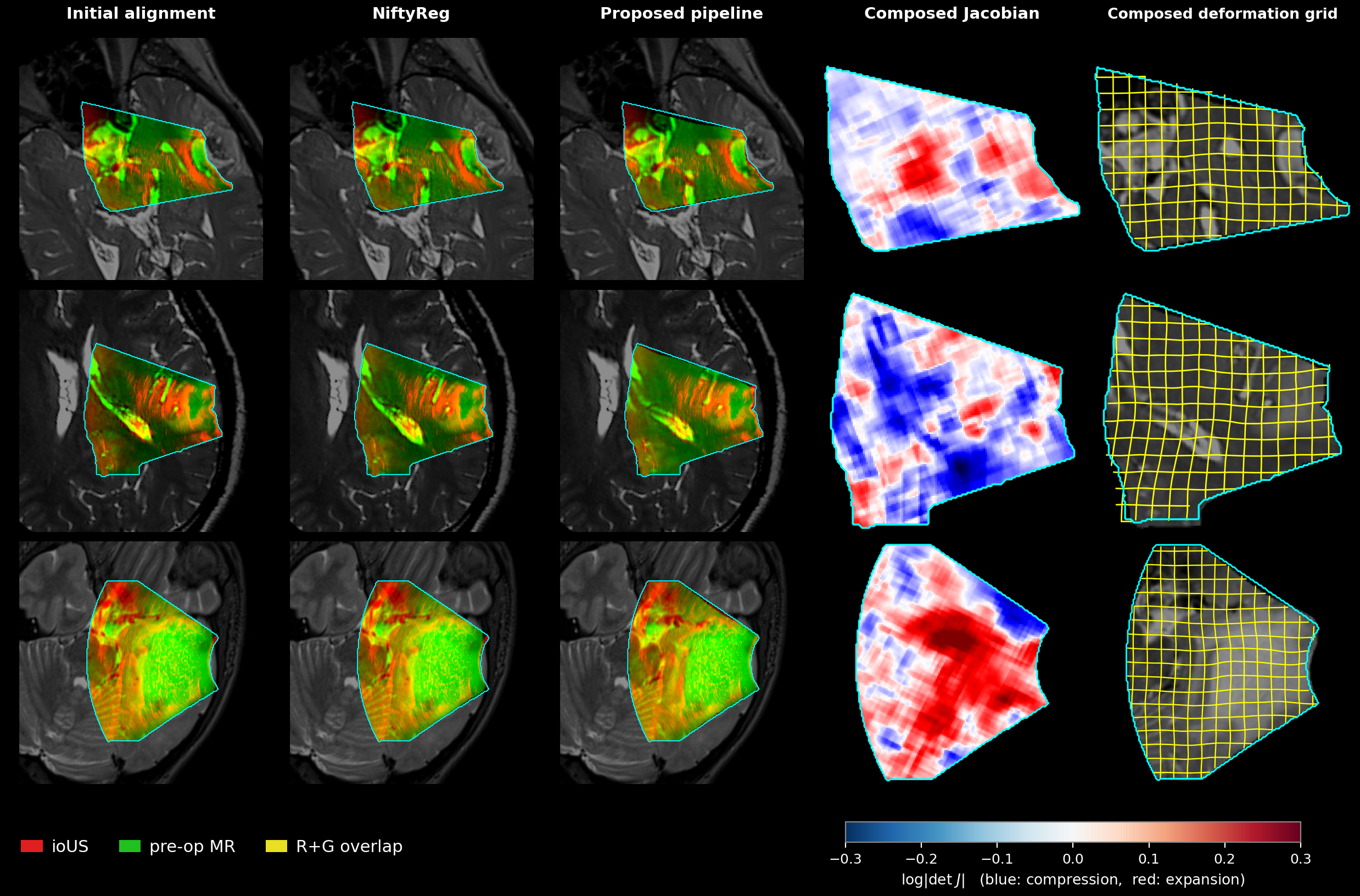}
  \caption{Geometry of the composed transformation for three representative
  subjects (rows). \textbf{Columns 1--3}: post-resection ioUS (red) overlaid
  with the preoperative MRI (green) inside the field-of-view cone; yellow marks
  voxels bright in both (good alignment), for the initial alignment, the
  NiftyReg classical baseline and the proposed pipeline. \textbf{Column 4}:
  log-Jacobian determinant $\log\,|\det J|$ of the composed transformation
  inside the cone (blue: compression; red: expansion). \textbf{Column 5}:
  composed deformation grid ($10$-voxel step) on the warped preoperative MRI.
  The cyan outline marks the ioUS field-of-view cone.}
  \label{fig:plausibility}
\end{figure}

We report Jacobian metrics on the composed transformation throughout, rather than on the SynthMorph stage in isolation, so that the reported geometric plausibility reflects the deployed pipeline rather than a property of an intermediate stage. Figure~\ref{fig:plausibility} illustrates the composed Jacobian and deformation grid for three representative subjects.

\subsubsection{Ablations of the synthesis-anchored stage}\label{sec:results-ablation}

We ablated the synthesis-anchored SynthMorph stage along three axes: the fixed-image weighting (uniform vs five segmentation-aware variants), the inference schedule (single-scale vs three-scale composed cascade vs three-scale weighted fusion) and the synthesis backbone (ResViT-2.5D vs ResViT-3D). Table~\ref{tab:ablation} gives the headline numbers; the full pairwise paired-Wilcoxon test matrix is provided in the supplementary material.

\begin{table}[t]
\centering
\caption{Ablation summary of the synthesis-anchored stage on $\TRE$. Paired Wilcoxon over the test cohort. Mean $\Delta\TRE$ is the per-subject mean of the proposed-minus-comparison difference; positive values favour the reference (the proposed configuration when the comparison is against an ablation).}
\label{tab:ablation}
\footnotesize
\setlength{\tabcolsep}{4pt}
\begin{tabular}{llcl}
\toprule
Design axis & Comparison & Mean $\Delta\TRE$ (mm) & $p$ \\
\midrule
Fixed-image weighting & uniform vs noCavity         & $-0.002$ & 0.39 \\
Fixed-image weighting & uniform vs noCavityNoTumor  & $+0.013$ & 0.49 \\
Fixed-image weighting & uniform vs tumorEmphasis    & $-0.016$ & 0.23 \\
Fixed-image weighting & uniform vs weightedDecay    & $-0.051$ & 0.12 \\
Inference schedule    & single-scale vs composed cascade & $+1.62$ & $4 \times 10^{-3}$ \\
Inference schedule    & single-scale vs weighted-fusion  & $+0.19$ & 0.03 \\
Synthesis backbone    & ResViT-2.5D vs ResViT-3D    & $+0.07$ & 0.54 \\
Rigid-only initialisation & vs vs proposed pipeline (ResViT-2.5D)   & $+0.43$ & $\mathbf{2 \times 10^{-3}}$ \\
\bottomrule
\end{tabular}
\end{table}

Four sub-findings emerge. First, fixed-image segmentation-aware weighting (zeroing the cavity, retaining only parenchyma, emphasising the tumour, or applying an exponential decay from the cavity boundary) provides no measurable benefit over uniform weighting. Inspection of per-subject deltas reveals that the cavity and tumour together occupy fewer than $5\%$ of FOV voxels in most subjects, and the SynthMorph implicit regularisation already absorbs any residual mismatch in that small region. This is the empirical justification for the segmentation-free deployment configuration of Section~\ref{sec:reg-proposed}: the segmentation-aware variants do not add value over uniform weighting, so the deployed pipeline drops the dependency on tumour or cavity labels altogether. Second, the composed cascade degrades $\TRE$ by approximately $1.6$~mm and visibly destroys the field plausibility. The single-scale schedule is robust by a wide margin. Third, the choice between ResViT-2.5D and ResViT-3D as the synthesis backbone of the proposed pipeline does not change the $\TRE$ outcome ($\Delta = +0.07$~mm, $p = 0.54$); the synthesis-quality and operational arguments of Sections~\ref{sec:results-synth-quality} and \ref{sec:backbone-selection} are the decisive ones. Fourth, an isolated SynthMorph deformable stage initialised by rigid-only \reg{reg\_aladin} (without the f3d MIND-SSC step) does \emph{not} improve over Initial ($\Delta = +0.03$~mm, $p = 0.67$). The NiftyReg \reg{reg\_f3d} stage is the load-bearing component of the proposed pipeline; the SynthMorph residual on top of it acts as a fine MRI-versus-MRI refinement, not as a replacement for the classical deformable solver.

\subsubsection{Per-subject improvement and near/far stratification}\label{sec:results-strat}

The mean $\TRE$ of $5.86$~mm of the proposed pipeline on our cohort is dominated by a handful of subjects with large residual brain shift; the typical per-subject improvement over Initial is informative. Across the 14 test subjects, the proposed pipeline reduces $\TRE$ relative to Initial in $10/14$ subjects (median reduction $0.39$~mm); the four subjects that do not benefit have Initial $\TRE$ at or below the present cohort median ($\sim 5.7$~mm), where the residual brain-shift signal is weakest. The classical baseline yields a similar $11/14$ improvement rate with a comparable median reduction.

Stratifying the per-pair $\TRE$ by Euclidean distance to the cavity-or-tumour segmentation (threshold $15$~mm) gives the near/far columns of Table~\ref{tab:reg-main}. On the Initial condition $\TRE$-near is $5.81$~mm versus $\TRE$-far $6.14$~mm: landmarks further from the surgical opening are harder on average than peri-cavity landmarks. Landmarks far from the cavity tend to lie more peripherally in the cube, where the rotation component of the residual pose contributes a larger displacement in millimetres; the peri-cavity brain-shift component is smaller in absolute terms and overlaid on this peripheral rigid error.

All registration methods preserve the same near $<$ far ordering. The classical NiftyReg baseline reduces $\TRE$-near to $5.25$~mm and $\TRE$-far to $5.80$~mm (a $0.56$~mm reduction near and $0.34$~mm reduction far over Initial); the proposed pipeline reduces $\TRE$-near to $5.39$~mm and $\TRE$-far to $5.80$~mm ($0.42 / 0.34$~mm reductions). The two methods are statistically indistinguishable on both strata.

\subsection{Synthetic intraoperative MRI}\label{sec:results-pseudo}

\begin{figure}[p]
\centering
\includegraphics[width=\textwidth,height=0.9\textheight,keepaspectratio]{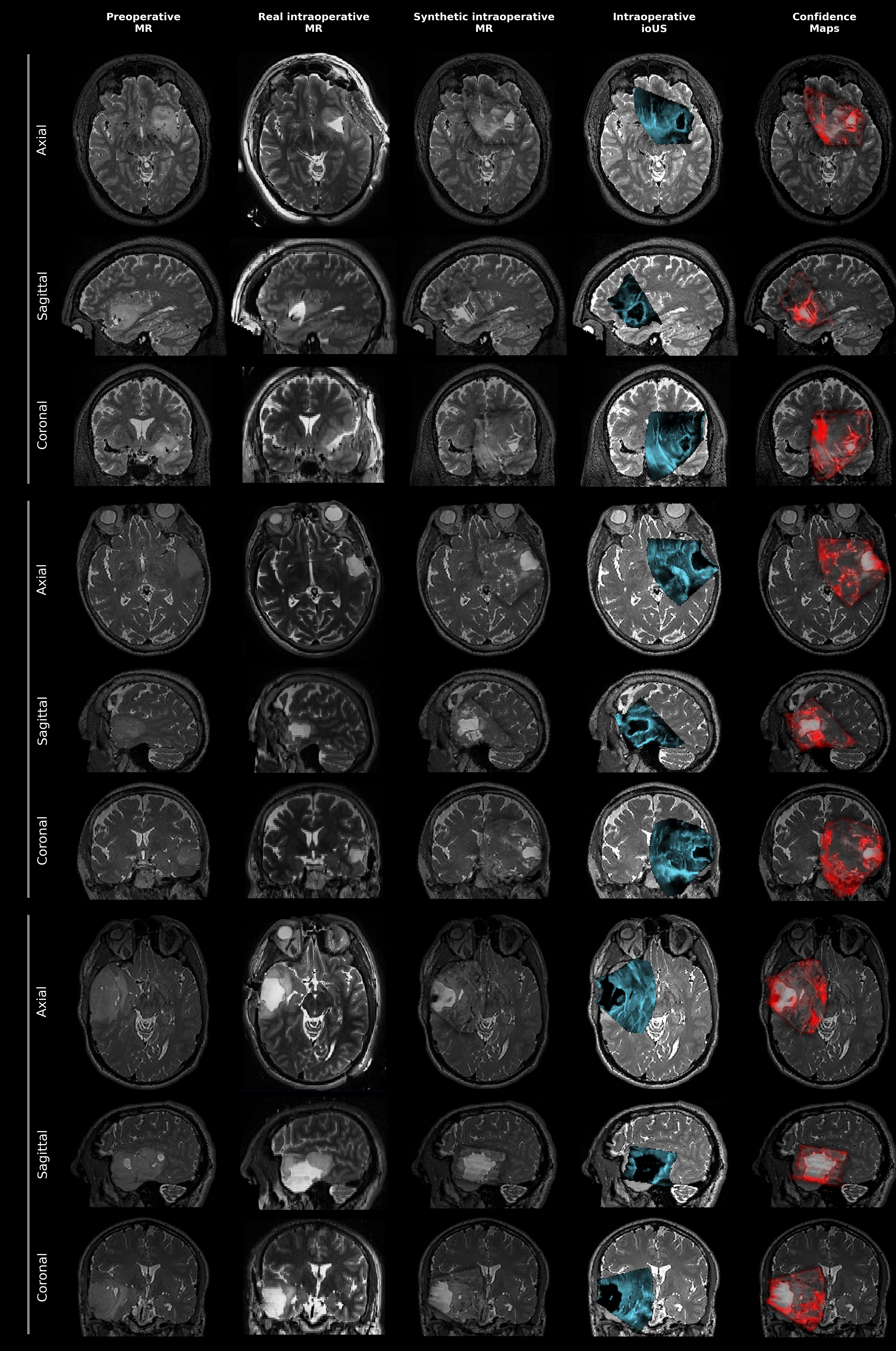}
\caption{Synthetic intraoperative MRI (siMRI) for three representative
subjects, each shown in three orthogonal planes (rows: axial, sagittal,
coronal). Columns from left to right: warped preoperative MR; real
intraoperative (post-resection) MR reference; synthetic intraoperative MR
(in-FOV deliverable); the ultrasound-insert variant with the intraoperative
ioUS field overlaid in cyan; and the per-voxel confidence map, in which
\textbf{red} denotes lower test-time-augmentation confidence.}
\label{fig:pseudo-mr}
\end{figure}

The siMRI is the deliverable of the pipeline and is produced for every subject of the test cohort in two variants (synthesis and ultrasound-insert, Section~\ref{sec:pseudo-cube}) and at two resolutions (intraoperative cube and original preoperative MRI grid, Section~\ref{sec:wholebrain}). Figure~\ref{fig:pseudo-mr}  shows three representative subjects covering different brain-shift magnitudes and tumour locations, each in axial, sagittal and coronal views, alongside the warped preoperative MR, the real intraoperative MR reference, the synthetic intraoperative MR, the ultrasound-insert variant and the per-voxel confidence map.

\subsubsection{Qualitative observations}\label{sec:results-pseudo-qual}

Three properties are visible across subjects. First, the synthesis content fills the FOV cone with MRI-like appearance that aligns spatially with the ioUS gyri, ventricular margins and falx midline, and the $\sigma = 5$~mm FOV feathering blends it into the warped preoperative MRI without a hard intensity seam. Second, inside the resection cavity the synthesis content recovers the cavity appearance from the ioUS, so the composed image reflects the post-resection anatomy rather than the intact preoperative tissue. Third, the whole-brain back-projection in the patient's preoperative imaging space places the surgical-opening contents at the anatomically correct location. The cone-projection boundary itself remains discernible, which is intended rather than a limitation: it marks for the surgeon the region covered by the ioUS, and therefore the part of the volume that has been updated to the intraoperative state, while the $\sigma = 5$~mm feathering keeps that boundary free of a hard intensity step.

\subsubsection{Coherence checks}\label{sec:results-pseudo-qc}

In addition to the absolute-quality comparison between the synthesis output and the intraoperative T2 inside the FOV cone (Section~\ref{sec:results-synth-quality}, Table~\ref{tab:synth-quality}), we report three coherence checks that operate on the cube-frame siMRI composition itself.

The first is the field plausibility of the composed transformation (Section~\ref{sec:results-plausibility}), already reported in Table~\ref{tab:reg-main}: composed folding $0.000\%$, composed SDLogJ $0.080$, median $|J|$ inside FOV close to 1.

The second is the normalised cross-correlation between the ioUS and the synthesis content inside the cube, $\NCC(\text{ioUS}, S)$, which is expected to be moderate: a value near zero would indicate a synthesis output that has lost spatial correspondence with the ioUS, whereas a value near one would indicate an output that has merely copied the ioUS speckle instead of producing MRI-like content. On the 14-subject cohort we observe $\NCC(\text{ioUS}, S)$ between $0.08$ and $0.54$ (median $0.33$, mean $0.32$), consistent with a synthesis output that preserves anatomical correspondence without reproducing speckle. The few subjects at the low end are those with the largest residual brain shift, and correspond to the highest landmark $\TRE$ values in Table~\ref{tab:reg-main}.

The third is the $\NCC$ between the ioUS and the ultrasound-insert variant $P_{\text{USinsert}}$, which by construction should approach unity inside the FOV: we observe $\NCC(\text{ioUS}, P_{\text{USinsert}}) = 1.000$ on $14/14$ subjects, confirming that the histogram-matching step and the FOV-feathered boundary are well behaved.

\subsection{Synthesis-confidence validation}\label{sec:results-confidence}

We validated the confidence map of Section~\ref{sec:confidence} as a predictor of synthesis error on the same 14-subject cohort and against the same reference used for the synthesis-quality benchmark (Section~\ref{sec:results-synth-quality}): the per-voxel absolute error $|S(\mathbf{r}) - T2_{\text{intraop}}(\mathbf{r})|$ between the in-FOV synthesis output, which under the deployed composition is the in-cone deliverable, and the raw intraoperative T2 resampled to the cube, both normalised by the $1$st and $99$th percentiles of the in-FOV reference foreground. A useful confidence map should assign low confidence where this error is large.

We report two error-prediction metrics. The \emph{sparsification error} progressively removes the least-confident voxels and tracks the mean error of those retained; its area above the oracle ordering (least-confident equals highest-error) is the area-under-the-sparsification-error, $\AUSE$ (lower is better), following Ilg et al. \citep{ilg_uncertainty_2018}. The \emph{Spearman} rank correlation $\rho$ between $-c(\mathbf{r})$ and the per-voxel error (subsampled to $2\times10^5$ voxels per subject) is positive when high confidence predicts low error. Both are computed per subject and aggregated as paired two-sided Wilcoxon signed-rank tests. We evaluated the deployed TTA confidence and the input-quality disclosure flag $s_{\text{input}}$ alone.

\begin{table}[t]
\centering
\caption{Synthesis-confidence validation against per-voxel synthesis error on the 14-subject test cohort. $\AUSE$ lower is better; Spearman $\rho$ between $-$confidence and error, higher (more positive) is better. Per-subject mean $\pm$ standard deviation. The deployed TTA-only map is in bold.}
\label{tab:confidence-val}
\footnotesize
\setlength{\tabcolsep}{6pt}
\begin{tabular}{lcc}
\toprule
Confidence variant & $\AUSE \downarrow$ & Spearman $\rho \uparrow$ \\
\midrule
\textbf{TTA only (deployed)} & $\mathbf{0.090 \pm 0.029}$ & $\mathbf{+0.200 \pm 0.107}$ \\
$s_{\text{input}}$ only & $0.139 \pm 0.031$ & $-0.101 \pm 0.092$ \\
\bottomrule
\end{tabular}
\end{table}

Table~\ref{tab:confidence-val} establishes two findings. First, the TTA confidence is the only variant whose correlation with synthesis error is positive; it attains the lowest $\AUSE$ ($0.090$) and the highest Spearman correlation ($+0.200$). The correlation is statistically positive but modest in magnitude, consistent with a map that flags the failure modes it is designed for rather than a calibrated error estimate. Second, the input-quality flag does not predict pixel-error magnitude in isolation: $s_{\text{input}}$ has a negative Spearman correlation with error ($-0.101$). It is therefore kept as a separate disclosure channel rather than aggregated into the confidence map, since folding an error-uncorrelated signal into the TTA score would only dilute it.

The negative result for the disclosure flag is expected and does not indicate that the flag is wrong. It flags regions where the ioUS signal is absent (acoustic shadow, cone edge); in exactly those regions the resampled intraoperative T2 reference is itself near zero, because the anisotropic native spacing of the reference ($0.86 \times 0.86 \times 2$~mm) spreads dark voxels under resampling, so the measured pixel error is artefactually small. The error-prediction framework reads this as a flag firing where the error is low, hence the negative correlation. We note that this reference artefact is load-bearing in more than one place (it also motivates restricting the similarity evaluation to the FOV cone in Section~\ref{sec:results-pseudo}), so it is stated as a caveat rather than relied upon: the TTA deliverable wins $\AUSE$, Spearman and the paired Wilcoxon comparisons regardless of how the reference is constructed, and the artefact only justifies retaining $s_{\text{input}}$ as a separate channel rather than discarding it.

The flag is therefore validated by construction against its design intent rather than against pixel error. Defining ioUS-dark voxels as those with local mean below $0.20\times$ the in-FOV speckle median, $s_{\text{input}}$ recalls $0.96 \pm 0.03$ of them at a precision of $0.32 \pm 0.13$ (per-subject mean $\pm$ standard deviation across the 14-subject test cohort): it rarely misses a true ioUS-dark region, at the cost of occasionally flagging clean speckle, which is the intended behaviour for a conservative warning that errs toward over-disclosure. The flag thus executes its intended failure-mode detection and is retained as a separate disclosure channel rather than folded into the confidence map.

%% ============================================================================
\section{Discussion}\label{sec:discussion}
%% ============================================================================

We have presented an end-to-end pipeline that, from a single intraoperative ultrasound acquisition and a preoperative MRI (both ingested as raw scanner NIfTIs with no preprocessing pre-alignment), produces a whole-brain synthetic intraoperative MRI aligned with the patient's preoperative imaging space. The pipeline combines a 2.5D residual-transformer synthesis backbone with a two-stage classical-plus-learning registration formulation and a composition step. Its evaluation on a 14-subject ReMIND-derived test cohort (215 landmark pairs, raw post-resection ioUS available) supports three claims that we discuss below: the synthesis backbone is selected by quality against the intraoperative T2 and by operational suitability for streaming deployment, with spacing robustness comparable between the 2.5D and 3D variants on our cohort; the synthesis-anchored learning residual matches classical registration on landmark $\TRE$ rather than improving it; and the operationally meaningful deliverable is the whole-brain synthetic intraoperative MRI rather than a marginal improvement in landmark target registration error.

\subsection{Why a 2.5D synthesis backbone}\label{sec:discussion-25D}
Esteban-Sinovas et al. \citep{estebansinovas_inprep} identify ResViT-3D as the highest peak-quality configuration against the warped preoperative T2 of the training distribution, with an advantage over the 2.5D variant of approximately 2~dB of $\PSNR$, $0.02$ of $\SSIM$ and $0.02$ of $\LPIPS$ at the training spacing. Against the intraoperative T2 of the present 14-subject test cohort, however, the ordering inverts: ResViT-2.5D outperforms its full-3D counterpart on $\SSIM$ ($p = 0.009$), $\PSNR$ ($p = 0.005$) and $\MAE$ ($p = 0.003$), and is statistically indistinguishable on $\LPIPS$ ($p = 0.22$; Section~\ref{sec:results-synth-quality}). The reference is the driver: the warped preoperative T2 carries the preoperative anatomy, which the 3D variant matches more faithfully at the training spacing; the intraoperative T2 carries the post-resection cavity and the brain-shift deformation that the 3D variant was not trained to reproduce. The slice-by-slice inference of the 2.5D variant absorbs this geometric mismatch better in the present cohort.
The robustness experiment of Section~\ref{sec:results-spacing-shift} provides a nuanced picture. ResViT-2.5D absorbs in-plane spacing changes through its internal slice-level resize to $256 \times 256$, an architectural prior that is invariant to in-plane spacing by construction; ResViT-3D operates on voxel-native sliding 3D windows and is therefore sensitive to the entire spacing vector in mm. Starting from the raw scanner ioUS at $\sim 0.13 \times 0.13 \times 0.5$~mm, neither backbone starts at its training native and the sensitivity-index ordering becomes metric-dependent: ResViT-2.5D wins on SSIM and NCC, ResViT-3D wins on LPIPS and MAE, and the two tie on PSNR (Table~\ref{tab:slopes}). The architectural argument therefore holds qualitatively but not as a uniform quantitative advantage. On the downstream registration $\TRE$, the two backbones remain statistically indistinguishable ($p = 0.54$, Section~\ref{sec:results-ablation}); the deciding factor between them is inference cost, with the 2.5D variant being approximately 6$\times$ faster per subject and naturally suited to slice-by-slice streaming for real-time deployment. SynDiff-2.5D is competitive with the ResViT variants on absolute image-quality metrics; we drop it from the deployed pipeline on grounds of iterative-diffusion latency, not output quality.

\subsection{The synthesis bridge does not improve registration $\TRE$}\label{sec:discussion-bridge}

A recurring intuition in the recent MRI-ioUS registration literature is that synthesising an MRI-like image from the ioUS turns the inter-modality problem into a more tractable intra-modality one, and that classical similarity terms should therefore benefit from the substitution. Our Section~\ref{sec:results-ablation} control test rejects this hypothesis on the present cohort: replacing the ioUS by the synthetic MRI as the fixed image for the classical \reg{reg\_aladin}~+~\reg{reg\_f3d} (MIND-SSC) stage produces a paired-Wilcoxon $p = 0.30$ in favour of the ioUS-fixed baseline. The interpretation is mechanistic: MIND-SSC is, by construction, invariant to modality-specific monotonic intensity transformations \citep{heinrich_mind_2013}. The classical solver already extracts the modality-independent content of the ioUS; the synthetic MRI does not add information that MIND-SSC was missing. This is consistent with the modest improvement that synthesis-augmented variants achieve on the ReMIND2Reg leaderboard against the same classical baseline \citep{wang_unsupervised_nodate}.

The synthesis bridge does have a measurable role on the learning-based displacement-field stage. SynthMorph \citep{hoffmann_synthmorph_2022} is pre-trained on a large multi-modal MRI corpus with synthetic intensity augmentation, and its similarity prior is calibrated to MRI-versus-MRI pairs. Applied directly to the raw (ioUS, preoperative MRI) pair, it produces small-magnitude displacements that do not fully recover the bulk brain-shift deformation; the ablation in Section~\ref{sec:results-ablation} confirms this directly, a rigid+SynthMorph schedule on the present cube fails to improve over Initial ($p = 0.67$). Applied on top of the NiftyReg-warped preoperative MRI, however, SynthMorph returns a clean, smooth, MRI-versus-MRI residual displacement that produces the synthesis-consistent warped preoperative MRI that the siMRI composition of Section~\ref{sec:pseudo-mr} needs. The composed two-stage pipeline therefore reaches the same landmark $\TRE$ as the classical baseline (paired Wilcoxon $p = 0.76$), not a better one. These differences are best read against the precision of the landmark reference itself. The inter-observer variability for manual localisation of this class of landmarks is approximately $1.89$~mm in 3D ultrasound \citep{machado_nonrigid_2018,dorent_brain_2025}, and is expected to be larger for the present set, which spans the MRI-to-ultrasound modality gap and targets peri-cavity and tumour-boundary fiducials. The $0.02$~mm gap between the proposed and classical pipelines is two orders of magnitude below this figure, so the two formulations are practically indistinguishable on the landmark endpoint; a registration accuracy statistically indistinguishable from the manual landmark variability is a recognised outcome in MRI-ultrasound registration \citep[e.g.][]{kuklisovamurgasova_registration_2013}, who report a fetal MRI-ultrasound registration whose target registration error ($2.52$~mm) did not differ significantly from the intra-rater landmark placement error ($2.14$~mm, $p = 0.81$). The $0.41$~mm improvement of both pipelines over the Initial world-frame alignment, although smaller than the per-landmark localisation scatter, remains resolvable in the per-pair mean over the $215$ landmarks and is statistically significant ($p \le 0.04$); we therefore read it as a modest but genuine recovery of brain shift rather than a clinically large correction. The operational value of the synthesis bridge is located in the siMRI deliverable, not in the landmark metric.

A separate point concerns the deformation-field plausibility of the composed pipeline. The composed transformation of the proposed pipeline is fold-free on every one of the $14$ test subjects, with an SDLogJ of $0.080$, and the classical NiftyReg baseline is correspondingly smooth (SDLogJ $0.007$ on the composed warp). The non-zero SDLogJ of the proposed pipeline reflects the larger non-rigid residual produced by the synthesis-anchored SynthMorph stage, not a folding risk. We continue to report field-plausibility metrics on the composed transformation throughout, so that the reported geometric plausibility reflects the deployed pipeline rather than a property of an intermediate stage.

\subsection{The synthetic intraoperative MRI as the operational deliverable}\label{sec:discussion-pseudo}

A registration pipeline that delivers only a warp from the preoperative MRI to the intraoperative space cannot represent the new anatomy that has appeared during the procedure. The resection cavity has no correspondence in the preoperative MRI; the warp can only deform preoperative tissue, not generate a fluid-bright cavity where there was solid tumour before. The synthesis backbone of the present pipeline addresses this gap: it produces an MRI-like image inside the ioUS field of view that reflects the intraoperative tissue state. Combined with the warped preoperative MRI outside the field of view and a feathered transition at the cone boundary, the composition yields a whole-brain volume that captures the intraoperative anatomy as an MRI, the modality the neuronavigator and the surgical team are calibrated to interpret. This deliverable is what classical MRI-ioUS registration alone cannot produce. (Section~\ref{sec:discussion-literature}).

Because the in-FOV deliverable equals the synthesis output (Section~\ref{sec:pseudo-cube}), the controlled within-subject comparison between the synthesis content and the true intraoperative T2 reported in Section~\ref{sec:results-synth-quality} (Table~\ref{tab:synth-quality}) is a direct quantification of the in-cone deliverable, not of an upstream component only. The evidence is therefore one of perceptual and intensity fidelity to the intraoperative contrast rather than of voxelwise structural co-registration to the post-resection MRI, which would in any case be bounded by the residual brain-shift recovery error of the registration stage (Section~\ref{sec:results-tre}) rather than by the synthesis. A reader-rated visual evaluation by neurosurgeons on a prospective cohort would complement these quantitative endpoints and is part of our ongoing work.

A pure-synthesis in-FOV deliverable raises a legitimate safety concern: unlike a blend anchored to the patient's real preoperative anatomy, it carries no structural guardrail against a synthesis error, so the surgeon would be navigating on a fully synthetic image inside the field of view. The per-voxel confidence map of Section~\ref{sec:confidence} is the deployment-side answer to this concern. Rather than re-introducing the preoperative anatomy as a blend, which our composition analysis shows would dilute the intraoperative content the deliverable is meant to provide, we ship the synthesis output together with an explicit per-voxel signal of where to distrust it. The map is the only one of the candidate uncertainty signals that is positively correlated with synthesis error (Section~\ref{sec:results-confidence}), and it is deliberately conservative, biasing toward over-reporting uncertainty. Its correlation with error is modest and it does not detect plausible hallucinations (a synthesis value in the typical MR range that is also stable under symmetry), so it is an interpretable deployment aid rather than a calibrated error probability; we return to this in Section~\ref{sec:discussion-lim}.

\subsection{Comparison with the literature}\label{sec:discussion-literature}

The mean $\TRE$ of $5.85$~mm achieved by the strongest classical baseline on our test cohort is higher than the $2.87$~mm reported by the same NiftyReg pipeline on the ReMIND2Reg 2024 challenge release \citep{dorent_brain_2025,hansen_learn2reg_2025} and higher than the $2.08$ to $2.28$~mm reported by cDRAMMS \citep{machado_deformable_2019} on BITE, RESECT and MIBS. Three factors explain the gap. First, our test cohort uses \textbf{post-resection ioUS}, after the cavity has formed and the brain has shifted; the BITE, RESECT and CuRIOUS2018 \citep{xiao_evaluation_2020} cohorts use pre-durotomy ioUS, when the brain is still in the preoperative configuration and brain shift is minimal. The post-resection task is intrinsically harder. Second, the ReMIND2Reg leaderboards report results on a cohort whose ioUS has been rigidly co-registered to the MRI partner before distribution; our pipeline operates on raw scanner inputs and therefore has to recover that rigid offset as part of the registration. The recent ReMIND2Reg 2025 release \citep{dorent_brain_2025} reports an initial $\TRE$ of approximately $4.81$~mm and a top-performer $\TRE$ of $4.42$~mm on the pre-aligned data; our $5.85$~mm is the value of a pipeline that takes raw scanner inputs without any preprocessing pre-alignment. Third, our landmark set (Section~\ref{sec:tre}) deliberately includes peri-cavity and tumour-boundary fiducials, the hardest regions to register; aggregate $\TRE$ is therefore expected to be higher than on benchmarks that distribute landmarks uniformly over the brain. These three factors are additive rather than alternative: the best ReMIND2Reg 2025 entry reaches $4.42$~mm on pre-aligned data and still sits well above the $\sim 1.89$~mm inter-observer floor \citep{machado_nonrigid_2018,dorent_brain_2025}, so a substantial part of the residual reflects post-resection brain shift that no method fully recovers, while our raw-input, peri-cavity-weighted evaluation accumulates the remaining offset in an interpretable way. We therefore read the $5.85$~mm not as a registration failure but as the cost of a deployment-realistic setting reported against a deliberately hard landmark set.

\subsection{Limitations}\label{sec:discussion-lim}

Four limitations of the present work deserve explicit acknowledgement.

\textbf{Cohort size.} Fourteen subjects with 215 valid landmark pairs is the largest cohort available for which we could obtain both the synthesis training held-out subset and the landmark-rich registration evaluation under a single annotation protocol; it is also at the upper end of the typical sample size for MRI-ioUS registration evaluations in this domain. Nonetheless, paired Wilcoxon comparisons at $n = 14$ have limited power to detect sub-millimetre $\TRE$ differences. The negative results we report (e.g.\ mask-variant ablations, 2.5D vs 3D synthesis backbones for downstream $\TRE$) should be read as the absence of evidence for a strong effect rather than evidence of strict equivalence. Equally, the equivalence we report between the proposed and classical formulations is stated relative to an inter-observer annotation floor of roughly $1.89$~mm \citep{machado_nonrigid_2018,dorent_brain_2025}, below which the present landmark reference cannot resolve differences.

\textbf{Single MRI sequence.} The pipeline is currently evaluated on T2-weighted preoperative MRI only. The 2.5D inference architecture and the registration formulation are sequence-agnostic in principle; portability to other MRI sequences is left to future work (Section~\ref{sec:discussion-future}).

\textbf{Composed-pipeline field plausibility.} The composed transformation is fold-free on every one of the $14$ test subjects (Section~\ref{sec:results-plausibility}). SDLogJ is moderate ($0.080$ for the proposed pipeline, $0.007$ for the classical baseline); the SDLogJ gap reflects the larger non-rigid residual produced by the SynthMorph stage on top of the NiftyReg-warped MRI, not a folding risk.

\textbf{Confidence-map scope.} The synthesis-confidence map is an interpretable deployment aid, not a calibrated probability of error. Its correlation with per-voxel synthesis error, while the strongest among the candidate signals, is modest (Spearman $\rho = +0.200$; Section~\ref{sec:results-confidence}), and by construction it cannot detect a plausible hallucination: a synthesis value that falls in the typical MR intensity range and is stable across the $D_4$ test-time augmentations is reported as high-confidence even when it is wrong. Furthermore, the error-prediction validation is referenced against the intraoperative T2 resampled from anisotropic native spacing, which introduces directional blur and biases the reference-side metrics toward smoother predictions; the same artefact underlies the negative correlation of the $s_{\text{input}}$ disclosure flag discussed in Section~\ref{sec:results-confidence}. A prospective, reader-rated assessment of whether the confidence overlay changes intraoperative interpretation is the appropriate next validation and is left to future work.

\subsection{Future work}\label{sec:discussion-future}

Four extensions are natural from the present pipeline.

First, the synthesis backbone could be replaced by a real-time slice-streaming variant of ResViT-2.5D that pipelines synthesis with ioUS acquisition, providing the surgeon with a continuously updated siMRI rather than the per-volume reconstruction reported here. The slice-by-slice inference of the 2.5D architecture is naturally suited to this streaming setting; the per-slice latency on consumer GPUs is in the tens of milliseconds.

Second, the pipeline is currently driven by a single ioUS acquisition. The siMRI could be updated over the course of the procedure by repeated ioUS acquisitions, with the composition step folding the new ioUS into the existing siMRI rather than overwriting it. This temporal extension also opens the door to per-acquisition uncertainty quantification, for instance by running multiple seeds of the SynthMorph residual stage or by deploying SynDiff-2.5D as a stochastic sampler for the in-FOV content.

Third, the pipeline should be deployed and validated in the operating room during real glioma resection procedures. A prospective study with neurosurgeon-rated siMRI overlays will quantify the clinical utility of the deliverable beyond the offline metrics reported here.

Fourth, the present pipeline is restricted to the T2-weighted sequence for the reasons given in Section~\ref{sec:splits}. Extending the synthesis stage to additional preoperative MRI sequences (notably T2-FLAIR, where lesion boundaries are most conspicuous) is a natural next step. The technical infrastructure of the pipeline (cube preprocessing, registration formulation, composition step) is already sequence-agnostic; the synthesis network would need to be retrained on the corresponding paired studies.

%% ============================================================================
\section{Conclusion}\label{sec:conclusion}
%% ============================================================================
We have presented an end-to-end pipeline that turns a single intraoperative ultrasound acquisition and the patient's preoperative MRI into two co-registered views of the operative field: the intraoperative ultrasound aligned to the preoperative MRI, and a synthetic MRI that stands in for an intraoperative MRI of the resected region. The pipeline couples a 2.5D residual-transformer synthesis backbone, chosen for its fidelity to the intraoperative T2 and its suitability for streaming deployment, with a two-stage classical-plus-learning registration and a composition step that requires no tumour or cavity labels and no manual preparation beyond the alignment already provided by the neuronavigator. On a ReMIND-derived cohort of 14 subjects and 215 expert-placed landmarks, the registration accuracy of the proposed composition matches that of a strong classical baseline while yielding a transformation that is diffeomorphic in every subject; unlike that baseline, the pipeline also reconstructs intraoperative anatomy, including the resection cavity, that registration alone cannot represent. With a per-subject runtime of a few minutes on a single consumer GPU, the resulting whole-brain volume could provide the surgeon with an MRI-like update of the operative field without dedicated intraoperative MRI infrastructure, with the potential to be integrated into current surgical-navigation systems.

%==============================================================================
\section*{Acknowledgements}
%==============================================================================
The authors thank Mª Luz de Andrés Loste, documentalist librarian at Hospital Universitario Río Hortega, for her assistance with literature retrieval.
%==============================================================================
\section*{Funding}
%==============================================================================
This research received no specific grant from any funding agency in the public, commercial or not-for-profit sectors.
%==============================================================================
\section*{Conflicts of Interest}
%==============================================================================
The authors declare no competing interests.
%==============================================================================
\section*{Data Availability}
%==============================================================================
The code supporting this study will be made available in a public repository upon acceptance of the manuscript. The data can be requested from the corresponding author.
%==============================================================================
\section*{Generative AI Disclosure}
%==============================================================================
The authors used Claude (Anthropic) to correct English usage and improve readability; all AI-assisted text was subsequently reviewed and verified by the authors, who take full responsibility for the content. PaperBanana was used to generate some of the schematic figures.
%% ============================================================================
%% Bibliography (embedded to avoid arXiv .bbl/.bib conflicts)
%% ============================================================================

\end{document}